\documentclass[journal]{IEEEtran}
\usepackage{cite}
\usepackage{amsmath,amssymb,amsfonts}
\usepackage{algorithmic}
\usepackage{algorithm}
\usepackage{graphicx}
\usepackage{textcomp}
\usepackage{xcolor}
\usepackage[space]{grffile} % Allows for spaces in filenames in includegraphics tags
\usepackage{nidanfloat} % Places two-column floats at bottom of page.
\usepackage{float} % Forces [H] option for figures to be invoked.
\usepackage{arydshln}
\usepackage{subfig}
\usepackage{breqn}
\usepackage{csquotes}
\usepackage{booktabs}
\usepackage{multirow}
\usepackage{hyperref}

\usepackage{graphicx} 
\graphicspath{{figs/}}

\newcommand{\etal}{\textit{et al.}}

\makeatletter
\patchcmd{\@maketitle}
{\addvspace{0.5\baselineskip}\egroup}
{\addvspace{-1\baselineskip}\egroup}
{}
{}
\makeatother

\makeatletter
\newcommand{\linebreakand}{%
\end{@IEEEauthorhalign}
\hfill\mbox{}\par
\mbox{}\hfill\begin{@IEEEauthorhalign}
}

\begin{document}	
\bstctlcite{IEEEexample:BSTcontrol} % Indicated for shortening long authors lists.

\title{Iterative, Deep Synthetic Aperture Sonar \\ Image Segmentation}

\author{Yung-Chen Sun, Isaac D. Gerg, and Vishal Monga

\thanks{This work was supported by the Office of Naval Research via grant N00014-19-1-2513. \emph{(Corresponding author: Yung-Chen Sun)}}
\thanks{Yung-Chen Sun, Isaac D. Gerg, and Vishal Monga are with the School of EECS, Pennsylvania State University, University Park, PA 16802 USA (e-mail: yzs5463@psu.edu). Isaac D. Gerg is also with the Applied Research Laboratory, State College, PA 16801 USA. }
}
\maketitle
\thispagestyle{plain}
\pagestyle{plain}

% ==================================================================================================
% Contents
% ==================================================================================================
\begin{abstract}
Synthetic aperture sonar (SAS) systems produce high-resolution images of the seabed environment. Moreover, deep learning has demonstrated superior ability in finding robust features for automating imagery analysis. However, the success of deep learning is conditioned on having lots of labeled training data, but obtaining generous pixel-level annotations of SAS imagery is often practically infeasible.  This challenge has thus far limited the adoption of deep learning methods for SAS segmentation. Algorithms exist to segment SAS imagery in an unsupervised manner, but they lack the benefit of state-of-the-art learning methods and the results present significant room for improvement. In view of the above, we propose a new iterative algorithm for unsupervised SAS image segmentation combining superpixel formation, deep learning, and traditional clustering methods. We call our method Iterative Deep Unsupervised Segmentation (IDUS). IDUS is an unsupervised learning framework that can be divided into four main steps:  1) A deep network estimates class assignments. 2) Low-level image features from the deep network are clustered into superpixels. 3) Superpixels are clustered into class assignments (which we call pseudo-labels) using $k$-means. 4) Resulting pseudo-labels are used for loss backpropagation of the deep network prediction. These four steps are performed iteratively until convergence. A comparison of IDUS to current state-of-the-art methods on a realistic benchmark dataset for SAS image segmentation demonstrates the benefits of our proposal even as the IDUS incurs a much lower computational burden during inference (actual labeling of a test image). Because our design combines merits of classical superpixel methods with deep learning, practically we demonstrate a very significant benefit in terms of reduced selection bias, i.e. IDUS shows markedly improved robustness against the choice of training images. Finally, we also develop a semi-supervised (SS) extension of IDUS called IDSS and demonstrate experimentally that it can further enhance performance while outperforming supervised alternatives that exploit the same labeled training imagery. 
\end{abstract}

%\begin{IEEEkeywords}
%Seabed texture segmentation, superpixel, deep learning, unsupervised learning, synthetic aperture sonar (SAS).
%\end{IEEEkeywords}
\section{Introduction}
\begin{figure}[t]
    \centering
    \begin{tabular}{cc}
        \includegraphics[height=3.5cm]{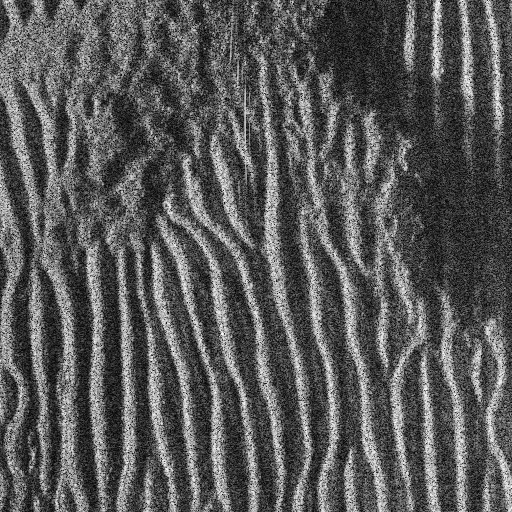}&\includegraphics[height=3.5cm]{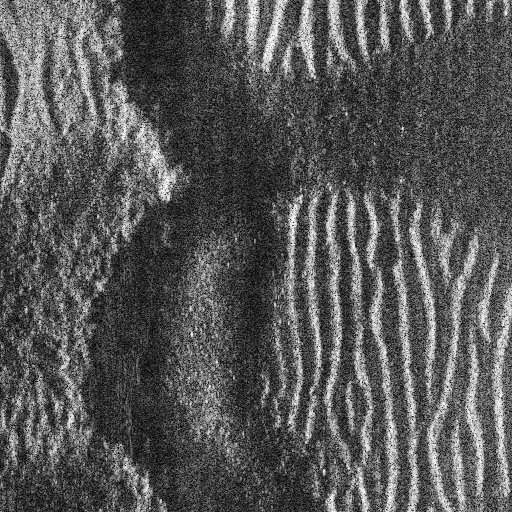}
    \end{tabular}
    \caption{Sample SAS images containing various seabed types including large sand ripples and flat sandy seafloor.}
    \label{fig:sonar_example}
 \end{figure}

SAS is a coherent imaging modality which is capable of producing high-resolution and constant-resolution images of the seafloor. This is an improvement over traditional side-scan-sonar (SSS) in which the resolution is function of range.  
Both theoretical and practical underpinnings of SAS imaging have been well established over the last few decades \cite{gough1986synthetic, hayes1991results, hayes1992broad, callow2003signal, fortune2001statistical, hunter2003simulation}. Currently, SAS systems deployed on autonomous underwater vehicles (AUVs) \cite{4099010} produce high-resolution images containing rich seabed information. The segmentation of these images plays an important role in analyzing the seabed environment. \figurename~\ref{fig:sonar_example} shows examples of SAS images containing complex seabed textures. Accurate segmentation of SAS imagery provides several benefits including large-scale environmental understanding of the seabed \cite{duval2016characterizing, trembanis2013detailed, davis2013distribution, denny2015use} and in-situ image interpretation to support AUV operations \cite{yordanova2019coverage}.

One challenge faced by SAS seabed segmentation is acquiring labeled data.  Obtaining pixel-level labels is time-consuming and often requires a diver survey to obtain correct results.  This lack of labeled data makes it difficult to deploy  convolutional neural networks (CNNs) for SAS seabed segmentation as such methods usually require lots of labeled data to work well. 

%In the past few years, deep learning frameworks applied for image segmentation have shown promising results \cite{liskowski2016segmenting, li2015cross, moeskops2016automatic}. Moreover, skip connections \cite{unet} between feature encoder and decoder perform exceptionally well in recognizing low-level motifs. Consequently, a naive intuition is that we can use similar network structures \cite{rahnemoonfar2019semantic,8734433} for seabed segmentation of SAS images because they are largely composed of low-level textures. However, training a convolutional neural network (CNN) often requires a considerable amount of labeled data to obtain good results \cite{krizhevsky2012imagenet}. For SAS image segmentation, this means acquiring pixel-level annotations which is a costly process and rarely do we have complete ground truth from a diver survey.  Consequently, humans only label regions which are easily recognized forgoing areas difficult to distinguish such as region boundaries or blended regions. As a result, few pixel-labeled datasets are available for training.

Many algorithms have been proposed to avoid the demand for labeled data \cite{Williams2015FastUS,Williams2010OnSR,Zare2017PossibilisticFL,Cobb2013MultiimageTS}. However, all of them are based on hand-crafted features which means they cannot jointly optimize feature extraction and clustering as the features are fixed. Some unsupervised image classification works \cite{caron2018deep,zhan2020online} have shown that embedding deep learning steps into a clustering algorithm is a promising direction.

%The essence of image-level clustering is to see every image as a region or ``big'' superpixel that only contains one high-level semantic texture, and the superpixel boundary is inaccurate. This idea provides a direction to generalize image-level clustering to pixel-level clustering. Although we are going to do pixel-level segmentation, from our observations, superpixels only containing a single texture will produce similar pixel predictions if their boundaries align textures well. Therefore, if we can find accurate boundaries to differentiate the regions (i.e., superpixels) containing a single texture, then the image segmentation process can be transferred into the superpixel classification problem.

In this work, we propose a CNN-based technique which overcomes the need to train on large datasets of pixel-level labeled data to achieve success. Specifically, we note the correlation between class boundaries of superpixels and human notated segmentation of SAS images\cite{cobb2014boundary}.  We devise a novel scheme which combines classical superpixel segmentation algorithms (which are non-differentiable) with deep learning (which is differentiable) to obtain a deep learning algorithm capable of using superpixel information of unlabeled pixels as a way to overcome the lack of abundant training data.

%The above intuition inspires us to devise a novel unsupervised segmentation method that combines deep learning and an iterative clustering technique producing an algorithm suitable for seabed segmentation of SAS images. More specifically, an IDUS iteration consists of learning feature representation and updating pseudo-labels. For learning features, it has no difference with the standard forward and backward propagation used in updating CNN's parameters; however, the labels for computing losses are automatically produced by the forward pass of the network along with unsupervised superpixel segmentation. For updating pseudo-labels, IDUS extracts pixel-level features (i.e., the feature map size equals the image size) from segmentation networks and generates the corresponding superpixels splitting one sonar image into many small regions. Each region is quantized into one superpixel feature and is the mean of the pixel features making up that superpixel. Then, a clustering process is performed to group these regions. In this way, our method iteratively updates CNN parameters, superpixels' boundaries, and feature centroids.

We evaluate our method, which we call iterative deep unsupervised segmentation (IDUS), against several comparison methods on a contemporary real-world SAS dataset.  We show compelling performance of IDUS against existing state-of-the-art methods for SAS image co-segmentation.  Additionally, we show an extension of our method which can obtain even better results when semi-labeled data is present.
%Besides this, the benefit of combining IDUS with supervised training is also be demonstrated to explore possible applications for the real-world case. 
\textbf{Specifically, our work makes the following contributions:}
\begin{enumerate}
    \item \textbf{We devise a new iterative method to integrate the merits of best-known unsupervised superpixel segmentation along with a deep feature extractor.} In particular, we utilize superpixels to generalize image-level clustering to superpixel clustering and propose an unsupervised learning framework for training a U-Net based CNN \cite{unet} in image segmentation tasks. Our proposal exceeds state-of-the-art performance on the task of unsupervised segmentation for seabed sonar images.
    \item We propose a supervised extension for IDUS, which called \textbf{Iterative Deep Semi-supervised Segmentation (IDSS)}, to increase network performance in situations that do not have enough annotated data.
    \item IDUS is \textbf{considerably faster} than the previously widely used methods, which encourages the deployment of IDUS on applications that require real-time speed.
    \item \textbf{Our approach does not depend on any specific network structure or superpixel generation algorithm.} The user can adjust the algorithm easily according to different segmentation tasks.
\end{enumerate}

\noindent \textbf{Reproducibility:} To facilitate verification of results reported in this paper as well as provide a resource for the SAS segmentation research community, we make our code and trained network available at: \url{https://scholarsphere.psu.edu/resources/ff521a5e-58e8-48b2-a9c9-5012800b62ab}
\section{Related Work}

\subsection{Seabed Texture Segmentation.}
Various unsupervised feature extraction methods related to our have been proposed. \cite{Williams2010OnSR} presents a sand ripple model that using sinusoid functions with different orientations and scales to characterize the sand ripple pattern as three highlight-shadow pairs. \cite{cobb2011autocorrelation} applies autocorrelation function to extract seabed features which perform especially well for extracting periodically patterns. However, it has been reported that autocorrelation is not a suitable measure for coarseness \cite{osti_6918260} (i.e., if a texture has bigger element sizes and fewer elements are repeated, the texture is coarser \cite{tamura1978textural}). Some works \cite{Cobb2013MultiimageTS,cobb2011autocorrelation} argue that single point statistics (such as gamma distribution variance or K-distribution shape parameter) are useful descriptors to represent seabed environments and easily differentiate between smooth and coarse textures. \cite{williams2009unsupervised} uses the coefficient of multi-scale orthogonal wavelet decomposition to extract the features where each sliding window of SAS images is treated as a unique data point. \cite{4730372, cobb2017multiple} use the superposition of Gabor filter banks response to represent seabed textures.

In traditional methods based on hand-designed features, the image patches usually are fed to a series of filter banks using sliding windows. Next, superpixels are used to quantize pixel-level features into regional features, and finally applied some clustering techniques such as $k$-means or Bayesian mixture model \cite{attias2000variational} to cluster these superpixels. Some works such as \cite{Cobb2013MultiimageTS, cobb2017multiple} would additionally use the idea of texton \cite{malik2001contour} to cluster the features into groups to find the prototypes of features before performing superpixel quantization. However, these kinds of hand-craft-based features are not differentiable and thus cannot be end-to-end trained. Therefore, the parameters of feature extraction are fixed and cannot be jointly optimized with the classifier.

There has been recent work in deep-networks \cite{rahnemoonfar2019semantic,8734433} trained for sidescan sonar images to segment seabed environment textures.  These models are trained in a purely supervised fashion (using tedious human annotation), with training and evaluation over a relatively homogeneous image collection. Hence, their generalization ability is difficult to ascertain.

\subsection{Unsupervised Deep Feature Learning}

Many works have been proposed for deep feature learning without any human-annotated labels and are generally summarized as self-supervised or clustering-based methods.

\subsubsection{Self-Supervised Methods}
For self-supervised learning, pretext tasks \cite{jing2020self} are pre-designed for networks to solve, and some characteristics of the data set are utilized to generate the pretext task goals (i.e., ground truth for pretext tasks) computing the loss function. In other words, visual features are learned by learning objective functions of pretext tasks. The commonly used pretext task includes generation-based methods such as image colorization \cite{zhang2016colorful}, image super-resolution \cite{ledig2017photo}, image inpainting \cite{pathak2016context}, image generation by Generative Adversarial Networks (GANs) \cite{10.5555/2969033.2969125,zhu2017unpaired}. Some pretext tasks hidden spatial information and let the CNN correctly restore it, includes image jigsaw puzzle \cite{noroozi2016unsupervised,wei2019iterative,ahsan2019video,kim2018learning}, context prediction \cite{li2016unsupervised}, and geometric transformation recognition \cite{zeiler2014visualizing,zhu2017unpaired}. Some pretext tasks are designed to learn visual features from videos providing rich temporal information. These tasks disorder the video frames first and force a CNN to predict if the input frame sequence is correct \cite{misra2016shuffle,wei2018learning} or the correct order of frame sequence \cite{lee2017unsupervised,xu2019self}. Although pretext-based methods can learn discriminative features, those approaches usually require experts that have the domain knowledge to carefully design. Therefore, it is not easy to transfer these approaches to other domains.

\subsubsection{Clustering-Based Methods}
The combination of clustering techniques and CNNs has recently been successful on small data sets in that it can go largely beyond the performance of  traditional frameworks, i.e., performs clustering without updating features \cite{xie2016unsupervised, yang2016joint, NIPS2014_07563a3f, faktor2014video}. For large-scale data sets such as ImageNet \cite{imagenet_cvpr09}, DeepCluster \cite{caron2018deep} provides an easily implemented solution using $k$-means to cluster the deep features and reassign new pseudo-labels to each image. This makes unsupervised learning of the features on large-scale datasets be possible. \cite{zhan2020online} solves the instability problem of DeepCluster by breaking down the global cluster into the batch-wise deep features clustering and labels update. These developments motivate us to develop our own work in unsupervised image segmentation wherein we will seek to integrate domain insight of unsupervised superpixel based sonar/SAS segmentation methods with the modeling power of deep learning.
\section{Method} \label{method}
% TODO: ref -> eqref for all the equation reference
\subsection{Background and Notation}

\begin{figure*}[t]
\setlength{\belowcaptionskip}{-0.5cm} 
    \centering
    \begin{tabular}{c}        
        \includegraphics[width=1.0\linewidth]{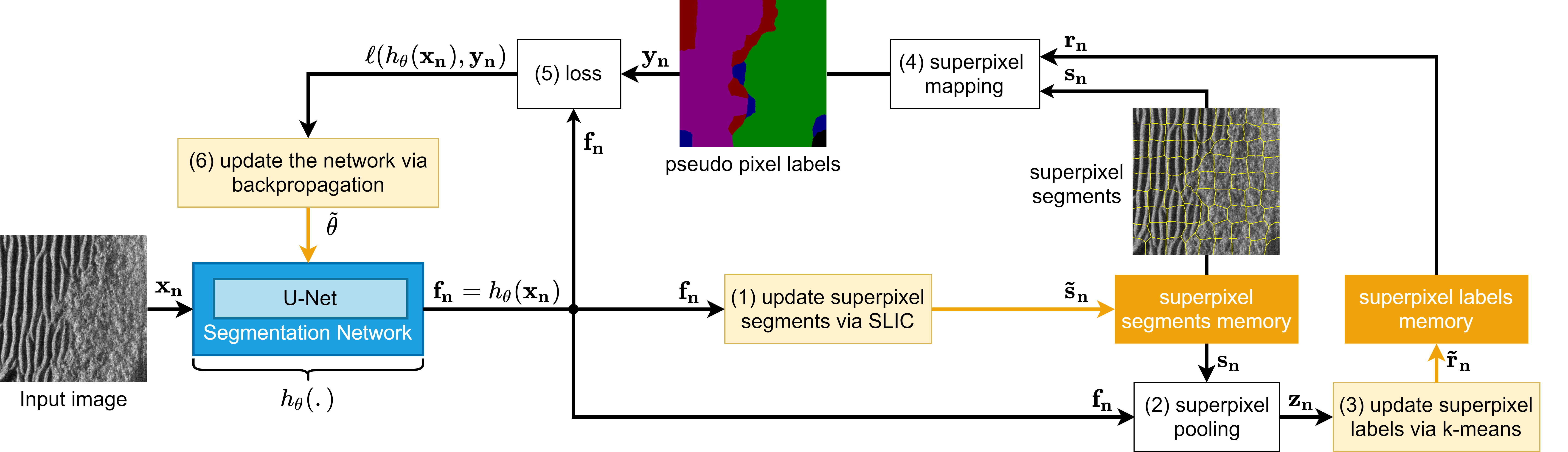}
    \end{tabular}
    \caption{
    \small
        The steps for an iteration of IDUS assuming there is only one input image: (1) The features are fed into SLIC\cite{SLIC} to generate new superpixel segments and replace the original ones in memory. (2) The pixel features are quantized into superpixel features with the new segments. (3) The superpixel features are clustered by $k$-means to update the class assignments in memory. (4) The superpixel labels are mapped to pseudo pixel labels via superpixel mapping with the new segments. (5) A loss is computed via the pseudo pixel labels and the network softmax outputs. (6) Finally, the network is updated via backpropagation according to the loss.
    }
    \label{fig:IDUS}
\end{figure*}

Given a set $\mathbf{X}=\left\{\mathbf{x_1},...,\mathbf{x_N}\right\}$ of $N$ SAS images, $s_n^k$ is the $k$-th superpixel of the $n$-th image and $r_n^k$ are the pseudo-label of this superpixel. We use $h_{\theta}(\cdot)$ to denote the softmax output of a segmentation network (i.e., deep network), where $\theta$ is the parameters we need to optimize. The output  has same size with the input image. Given an input image $\mathbf{x_n}$, the network maps the image into a feature map $\mathbf{f_n}=h_{\theta}(\mathbf{x_n})$ in which $f_n^i$ represent the feature of $i$-th pixel $x_n^i$ in this image.

\subsubsection{Superpixel Pooling}
To quantize the pixel-level features into region-level features $\mathbf{z_n}=Q_{pool}(\mathbf{x_n},\mathbf{f_n},\mathbf{s_n})$, we compute the feature's mean from all the pixels making up the  superpixel $z_n^k$ by the following equation: 
\begin{equation}\label{eq:1}
   z_n^k=\frac{\sum_{i=1}^{D}f_n^iI(x_n^i \in s_n^k)}{\sum_{i=1}^{D}I(x_n^i \in s_n^k)},
\end{equation}
where $D$ denotes the number of pixels in the input image. $I(x_n^i \in s_k^i)$ is an indicator function that is 1 if $x_n^i$ belongs to the superpixel $s_n^k$. The superpixel quantization makes sure that we can collect regional features of sufficient accuracy for the later clustering use.

\subsubsection{Superpixel Mapping} 
Suppose we have already known a superpixel label $r_n^k$, then every pixel label $y_n^i$ in this superpixel should be the same. This mapping is given by:
\begin{equation}\label{eq:2}
    y_n^i=\sum_{k=1}^{K_n}r_{n}^{k}I(x_n^i \in s_n^k),
\end{equation}
where $K_n$ denotes the number of superpixels in $n$th image. We use $\mathbf{y_n} = Q_{map}(\mathbf{x_n},\mathbf{r_n},\mathbf{s_n})$ to denote this process which we call superpixel mapping. It eliminates some classification errors and produce more robust class assignments for pixels.
\vspace{-3mm}
\subsection{Iterative Deep Unsupervised Segmentation (IDUS)}
% TODO: need to move to the method section ?
% As far as we know, The state-of-the-art methods for unsupervised sonar image segmentation are all based on hand-crafted features which are fixed and do not change during training. This is in opposition to IDUS in which we jointly learn the features and segmentation as part of our training method.

DeepCluster \cite{caron2018deep} alternates between updating the model's parameters and reassigning pseudo-labels by $k$-means to the input images. Every image is mapped to a deep feature; however, this process loses boundary information between different textures. In other words, the deep feature can only represent a region with a rough boundary which is not accurate enough for unsupervised image segmentation. Different from DeepCluster, we use the idea of a superpixel to make sure that we can use the deep features to represent regions with accurate boundaries and allow learning of a segmentation network without any human-annotated pixel-labels. To better understand the proposed idea, we introduce the simplest version of IDUS first and then a complete description will follow.

\subsubsection{The Simplest Version of IDUS}
\figurename~\ref{fig:IDUS} shows an iteration of the simplest version of IDUS after initialization (will be introduced in Section \ref{init}). Here, we assume that the dataset only contains one image. We store the superpixel segments and labels in the superpixel segments memory and superpixel labels memory, respectively, since we will not update the segments and labels every training epoch for the consideration of computation efficiency. We see the softmax vectors $\mathbf{f_n}$ of network outputs as a kind of feature representation due to it has the same size of the inputs and contains the distance information between a region texture and cluster centroids. Given the feature map (pixel features)  $\mathbf{f_n}$ ingesting from the input image via a segmentation network, an iteration of IDUS contains six main steps:
\begin{enumerate}
    \item The pixel features are fed into SLIC\cite{SLIC} to generate new superpixel segments and replace the original ones.
    \item The pixel features are quantized into superpixel features (by Eq \eqref{eq:1}) with the new segments.
    \item The superpixel features are clustered by $k$-means to update the class assignments in memory.
    \item The superpixel labels are mapped to pseudo pixel labels via superpixel mapping (Eq \eqref{eq:2}) with the new segments.
    \item  A loss is computed via the pseudo pixel labels and the network softmax outputs.
    \item The network is updated via backpropagation. 
\end{enumerate}
%As shown in \figurename~\ref{fig:IDUS}, IDUS alternates between generating pseudo pixel labels (Steps 1-4) and updating the network parameters (Steps 5, 6), which is similar to DeepCluster \cite{caron2018deep}. %\textcolor{blue}{i cant following this section well.}

\subsubsection{The Complete Version of IDUS} More precisely, given a set of training images $\mathbf{X_T} = \left\{\mathbf{x_n}\right\}_{n=1}^{N_{T}}$, we update the model's parameters $\theta$ by solving the following problem,
\begin{equation}\label{eq:3}
    \min_{\theta}\frac{1}{N_{T}}\sum_{n=1}^{N_{T}}\ell(h_{\theta}(\mathbf{x_n}), Q_{map}(\mathbf{x_n},\mathbf{r_n},\mathbf{s_n}) ),
\end{equation}
where $\ell$ is a loss function and $\mathbf{r_n}$, $\mathbf{s_n}$ will be initialized before performing the algorithm. After training the model for $U_{E}$ epochs, we extract the feature maps $\left\{\mathbf{f_n}\right\}_{n=1}^{N_T}$ of all training images based on the current model's parameters $\theta_{\omega}$ where $\omega$ is the number of iteration. Then, we pool the feature maps to regional features $\left\{\mathbf{z_n}\right\}_{n=1}^{N_T}$ and cluster them into $M$ groups based on the optimization goal
\begin{equation}\label{eq:4}
    \min_{C\in\mathbb{R}^{M\times d}}\sum_{m=1}^M\sum_{n=1}^N\sum_{k=1}^{K_n}\min_{r_n^k \in [1, M]}\|z_n^k-c_m\|I(r_n^k==m),
\end{equation}
where $c_m$ is the centroid feature of class $m$ and $d$ is the length of feature dimension. With this clustering criterion, the regional features will be clustered and used to produced new labels $\left\{\mathbf{r_n}\right\}_{n=1}^{N_T}$; and then, they will be assigned to the corresponding superpixels.

The overall idea of IDUS can be summarized in Algorithm \ref{alg:algorithm1}. We update the superpixel boundaries every $U_{S}$ epochs and the algorithm terminates after a fixed number of iterations. Note, despite updating the superpixels based on the intermediate representation of the network, the loss function is still differentiable with respect to the input and network parameters. The network outputs, $\mathbf{f}_n$, are only needed during the back-propagation step (and hence not used during the forward pass) during training. Consequently, the superpixel clustering performed on $\mathbf{f}_n$ is implicitly learned during training while still allowing the objective function to be differentiable.  This is the same approach employed by the DeepCluster \cite{caron2018deep} algorithm on which our method is based.
%and as such, are free to take on whatever values they like while still maintaining differentiability of the objective function. This is no different than traditional neural network training, but now our target output value is also a function of our network output instead of being given as an input training point (often denoted as an $(X_i, Y_i)$ pair). 

\begin{algorithm}[t]
    \caption{Iterative Deep Unsupervised Segmentation}
    \begin{algorithmic}[1]
        \STATE Initialize superpixel segments, labels $\left\{\mathbf{s_n},\mathbf{r_n}\right\}_{n=1}^{N_T}$ of $N_T$ training images. And map the regional labels to pseudo pixel labels $\left\{\mathbf{y_n}\right\}_{n=1}^{N_T}$ by Eq \eqref{eq:2}.
        \STATE Assign intervals $U_{E}$, $U_{S}$ for updating the pseudo-ground-truth and superpixel boundaries.
        \STATE Set the iteration index $\omega = 0$ and assign a number for maximum iterations $\Omega$.
    \REPEAT
        \STATE Given training data set $\left\{\mathbf{x_n}, \mathbf{y_n}\right\}_{n=1}^{N_{T}}$, update the model's parameters $\theta_{\omega}$ based on the problem Eq~\eqref{eq:3}.
        
        \IF{every $U_S$ epochs}
            \STATE Update superpixel boundaries $\left\{\mathbf{s_n}\right\}_{n=1}^N$ based on the model of current iteration.
        \ENDIF
        \IF{every $U_E$ epochs}
            \STATE Use the model parameters of current iteration $\theta_{\omega}$ to generate pixel-level features $\{\mathbf{f_n}\}_{n=1}^{N_T}$.
            \STATE Pool the pixel-level features into regional features (superpixel features) by Eq \eqref{eq:1}.
            \STATE Cluster the regional features based on criterion Eq \eqref{eq:4} and reassign new pseudo labels $\left\{\mathbf{r_n}\right\}_{n=1}^{N_T}$ to each superpixel.
        \ENDIF
        \STATE $\omega\mathrel{+}=1$.
    \UNTIL{$\omega\geq\Omega$.}
    \STATE Output the model's parameters $\theta_{\Omega}$.
    \end{algorithmic}
    \label{alg:algorithm1}
\end{algorithm}

\subsection{Iterative Deep Semi-Supervised Segmentation (IDSS)}

\noindent Although pixel-level labels are expensive to collect, it is not uncommon to see the ``easy'' areas of the images labeled by human operators.  To utilize this extra information, we propose Iterative Deep Semi-Supervised Segmentation (IDSS) to combine IDUS with (semi-)supervised training and improve the network performance. Specifically, we use $\{\mathbf{X}_{unlabel}\}$ to denote the unlabeled data set and $\{\mathbf{X}_{label}, \mathbf{Y}_{label}\}$ the label data set where $ \mathbf{Y}_{label}$ is the corresponding ground truth. To combine IDUS with supervised training, we first use $\{\mathbf{X}_{unlabel}, \mathbf{X}_{label}\}$ to learn the visual features of U-Net by IDUS. Then, we extract the U-Net and re-initialize its segmentation head (linear classifier after the decoder). Finally, the pre-trained U-Net is fine-tuned on the labeled data set $\{\mathbf{X}_{label}, \mathbf{Y}_{label}\}$ which has fewer samples. We describe the exact design of IDSS in Algorithm \ref{alg:comb}.

    \begin{algorithm}[t]
        \begin{algorithmic}[1]
            \STATE Input the unlabeled dataset $\{\mathbf{X}_{unlabel}\}$ and labeled dataset $\{\mathbf{X}_{label},\mathbf{Y}_{label}\}$. 
            \STATE Use $\mathbf{X}_{unlabel}$ and $\mathbf{X}_{label}$ to learn feature representation by IDUS (Unsupervised Learning).
            \STATE Extract the backbone model U-Net from IDUS, and re-initialize its segmentation head.
            \STATE Use $\{\mathbf{X}_{label},\mathbf{Y}_{label}\}$ to fine-tune the U-Net (Supervised Learning).
            \STATE Output the model's parameters $\theta$ of the IDSS U-Net.
        \end{algorithmic}
    \caption{Iterative Deep Semi-Supervised Segmentation}
    \label{alg:comb}
    \end{algorithm}

\subsection{Implementation Details}\label{imp_dt}
\subsubsection{Initialization}\label{init}
We use the idea of texton \cite{malik2001contour} to initialize the pseudo labels and a two-step texton selection \cite{Cobb2013MultiimageTS} to reduce the computation time. We give a rough initialization to IDUS so that encourages the algorithm to converge. An initialization for IDUS contains the following steps: 
\begin{enumerate}
    \item We extract the feature maps of every image from the output of the first and third convolutional blocks in ResNet-18 \cite{7780459}, and the channel dimensions for two feature maps are PCA-reduced to 8 and 16 respectively.
    \item The feature maps from two layers are resized to $128\times128$ pixels and then concatenated together. Therefore, the new feature maps contain the information of high- and low-level semantics.
    \item We cluster each image’s features separately into 128 groups in order to find the local textons (i.e., each image has 128 centroid features). For the data set that has N images, we find $N \times 128$ local textons totally.
    \item These local textons are clustered into 128 groups again to find the global textons. And, the global textons are assigned to each pixel according to the distance measure between its features and the nearest global textons.
    \item The histogram of textons distribution around a pixel is computed within a $10 \times 10$ sliding window for smoothing purposes, where each histogram element represents the count of the corresponding global textons.
    Therefore, we get a $128 \times 128 \times 128$ histogram map and resize it to $512 \times 512 \times 128$ which has the same size as input images.
    \item The normalized histograms with range $[0, 1]$ are concatenated with the feature maps of Williams \cite{williams2009unsupervised} (the implementations will be described in Section \ref{up_co_seg}). We use these wavelet features because previous work \cite{williams2009unsupervised} has shown their utility in modeling SAS imagery.
    %been shown useful in SAS related features and could differentiate Seagrass and  Rock textures well}.
    \item We use SLIC \cite{SLIC} to generate 100 superpixels for each image based on the generated features and pooling the pixel-level features into superpixel histograms via Eq \eqref{eq:1}. And perform superpixel clustering via $k$-means to get the pseudo superpixel labels to drive the IDUS.
\end{enumerate}

To clarify the relationship between the initialization and IDUS, the initialization for IDUS is performed at $\textit{iteration 0}$ while $\textit{iteration 1},~\textit{iteration 2},~..., etc.$ are subsequent iterations. The only difference between $\textit{iteration 0}$ and the subsequent iterations is that we use a set of initial features $\mathbf{F_0}=\left\{\mathbf{f_n}\right\}_{n=1}^{N_T}$ generated by $steps~1-6$ to replace the segmentation network softmax outputs $\left\{\mathbf{f_n}\right\}$ as shown in \figurename~\ref{fig:IDUS}. $Step~7$ is the same procedure as $steps~1-3$ in \figurename~\ref{fig:IDUS}.

\begin{table}
    \centering
    \begin{tabular}{ccccc}
    \toprule
    Layer Name & Layer Function & Dim. & \# Filters & Input \\\midrule
    up1        & Upsampling     & 2$\times$2   & N/A      & res4 \\
    merge1     & Concatenate    & N/A          & N/A      & up1, res3 \\
    conv1a     & Conv+BN+ReLU    & 3$\times$3   & 256      & merge1 \\
    conv1b     & Conv+BN+ReLU     & 3$\times$3   & 256      & conv1a \\\midrule
    
    up2        & Upsampling     & 2$\times$2   & N/A      & conv1b\\
    merge2     & Concatenate    & N/A          & N/A      & up2, res2 \\
    conv2a     & Conv+BN+ReLU     & 3$\times$3   & 128      & merge2 \\
    conv2b     & Conv+BN+ReLU     & 3$\times$3   & 128      & conv2a \\\midrule
    
    up3        & Upsampling     & 2$\times$2   & N/A      & conv2b\\
    merge3     & Concatenate    & N/A          & N/A      & up3, res1 \\
    conv3a     & Conv+BN+ReLU     & 3$\times$3   & 64       & merge3 \\
    conv3b     & Conv+BN+ReLU     & 3$\times$3   & 64       & conv3a \\\midrule
    
    up4        & Upsampling     & 2$\times$2   & N/A      & conv3b\\
    merge4     & Concatenate    & N/A          & N/A      & up4, res\_conv \\
    conv4a     & Conv+BN+ReLU     & 3$\times$3   & 32       & merge4 \\
    conv4b     & Conv+BN+ReLU     & 3$\times$3   & 32       & conv4a \\\midrule
    
    up5        & Upsampling     & 2$\times$2   & N/A      & conv4b\\
    conv5a     & Conv+BN+ReLU     & 3$\times$3   & 16       & up5 \\
    conv5b     & Conv+BN+ReLU     & 3$\times$3   & 16       & conv5a \\\midrule
    
    seg\_head   & Conv+Softmax     & 3$\times$3   & 7        & conv5b \\
    \bottomrule
    \end{tabular}
\caption{Description of decoder and segmentation head of the U-Net used in IDUS.  This decoder is preceded by a pre-trained ResNet-18 network.  ``Conv+BN+ReLU" is 2D convolutional followed by Batch Normalization followed by rectified linear unit (ReLU) activation and ``Softmax" denotes the softmax function222. We use layers res\_conv, res1, res2, res3, and res4 to denote the output of ResNet-18's first convolutional layer and building blocks, respectively. Layer conv5b is the decoder output and layer seg\_head is the segmentation head output.} 
\label{tab:network_struct}
\end{table}

\subsubsection{Network Structure} 
The backbone model used by IDUS and IDSS is U-Net\cite{unet} equipped with ResNet-18 encoder that pre-trained on ImageNet \cite{imagenet_cvpr09}. \tablename~\ref{tab:network_struct} shows the decoder and segmentation head (pixel-level linear classifier) of U-Net. We use \{res\_conv, res1, res2, res3, res4\} to denote the output of ResNet-18's first convolutional layer and building blocks, respectively. The decoder output (conv5a) will be fed to the segmentation head (seg\_head) to generate the final network outputs. The decoder and segmentation head parameters will be re-initialized right after updating pseudo-labels since we find that doing this can improve the network performance.
%The reason might be that the pseudo-labels are not all correct and these inaccurate labels may induce the network to converge to the bad saddle point which will influence the training of the next iterations.

\subsubsection{Training Configuration for IDUS}\label{train_config}
We train the IDUS with mini-batch size fifteen on an NVIDIA Titan X GPU (12GB) with the PyTorch package \cite{NEURIPS2019_bdbca288}. The loss function we use is the mean of dice loss \cite{sudre2017generalised} and cross-entropy loss. We use weights $w_m^{ept} = 1/r_m$ and $w_m^{dice} = 1/\sqrt{r_m}$, where $r_m$ was the proportion of $m$th class label in training samples, to balance the class weights of two losses, respectively. Batch normalization layers \cite{ioffe2015batch} are added after each convolutional layer (except the segmentation head) to accelerate the loss convergence. The model's parameters are initialed by a uniform distribution following the scaling of \cite{he2015delving} and optimized by Adam \cite{kingma2014adam} algorithm with weight decay value $10^{-9}$. In every iteration, the learning rate of the model is started with $10^{-4}$ and decreased by a drop rate $0.1$ every 100 epochs. The updating interval of superpixel labels and boundaries is 200 and we do this five times (five iterations) throughout training. Consequently, we train for 1000 epochs where every 200 epochs we update the superpixel labels and boundaries.

\subsubsection{Training Configurations for IDSS}
In IDSS, the training configuration of unsupervised learning is the same as that of IDUS. For the supervised part, the learning rate was set to $10^{-4}$ and decreased by $0.1$ every eighty epochs. We first update the parameters of the decoder and segmentation head. After forty epochs, we add the encoder parameters to the optimizer to jointly optimize.
\section{Experiments and Results}
In this section, we describe how we measure the performance of IDUS and demonstrates its efficacy against the comparison methods. Besides this, the extension IDSS will also be evaluated to show the benefits of semi-supervised training. Section \ref{data} describes the SAS dataset we used and the associated preprocessing performed. Section \ref{converge_ana} analyzes the convergence of IDUS. Section \ref{up_co_seg} presents co-segmentation results. Section \ref{feature_quality} examines the quality of the unsupervised learned features. Section \ref{compute_pfm} presents the run time of each method. Section \ref{idssevl} presents the semi-supervised training results of IDSS when only some pixel-level labels are present. Finally, Section \ref{visual}, presents selected results of performing IDUS in a unsupervised fashion comparing its results to a semi-supervised trained U-Net, a popular deep learning architecture used for segmentation. 

\subsection{Dataset Description and Pre-processing}\label{data}
\begin{figure*}[t]
        \centering
        \begin{tabular}{cccc}
            \includegraphics[width=0.18\linewidth]{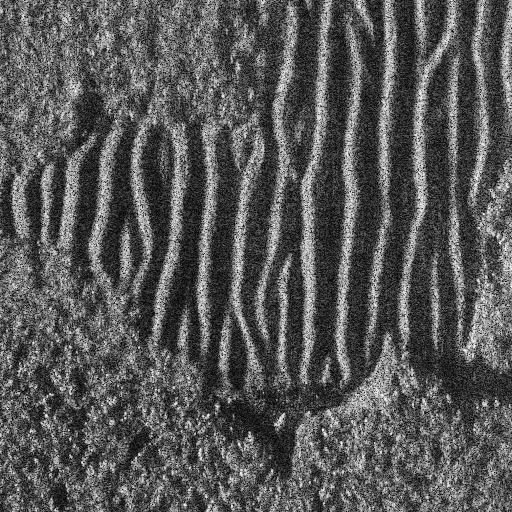}& \includegraphics[width=0.18\linewidth]{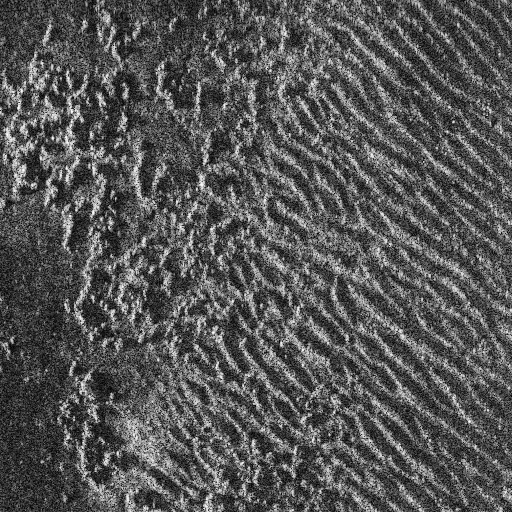}&
            \includegraphics[width=0.18\linewidth]{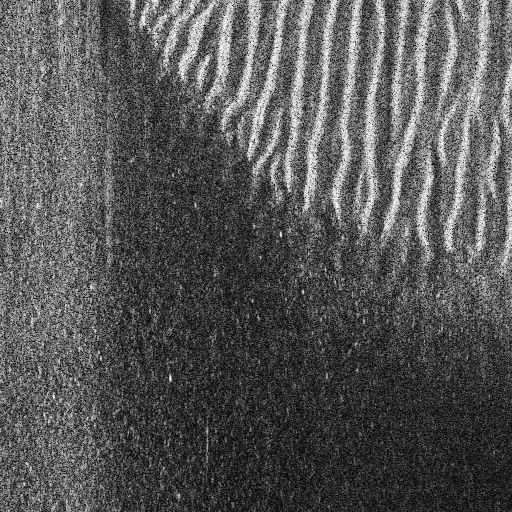}&
            \includegraphics[width=0.18\linewidth]{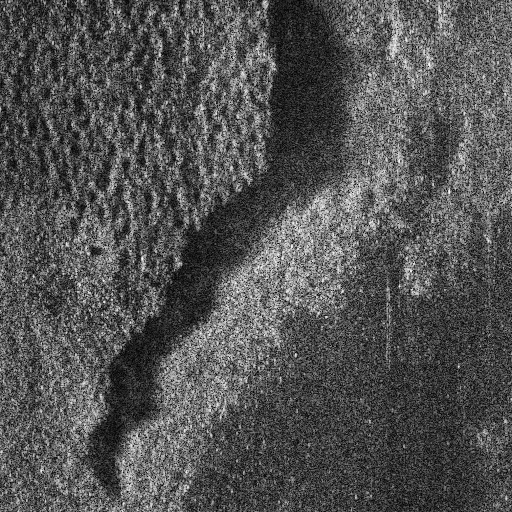}\\
            \includegraphics[width=0.18\linewidth]{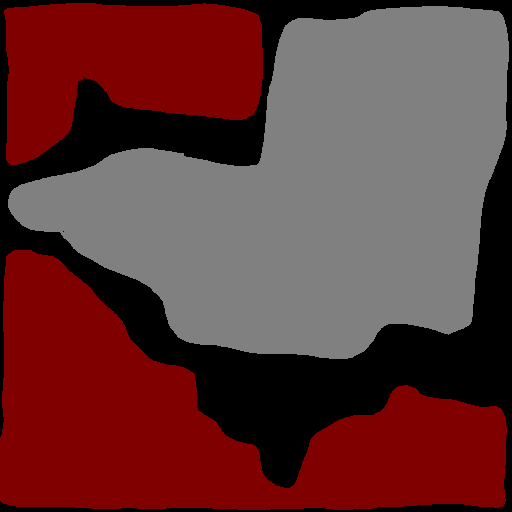}& \includegraphics[width=0.18\linewidth]{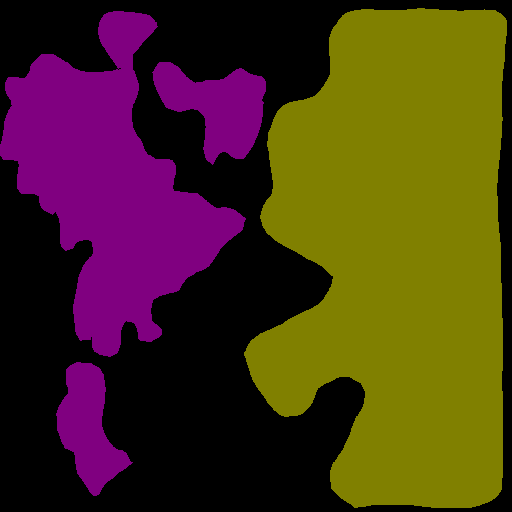}&
            \includegraphics[width=0.18\linewidth]{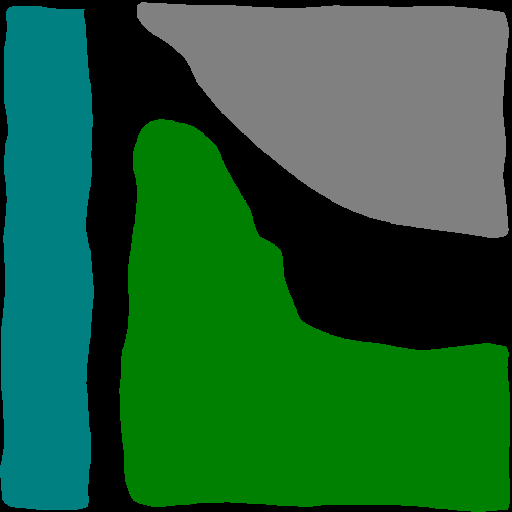}&
            \includegraphics[width=0.18\linewidth]{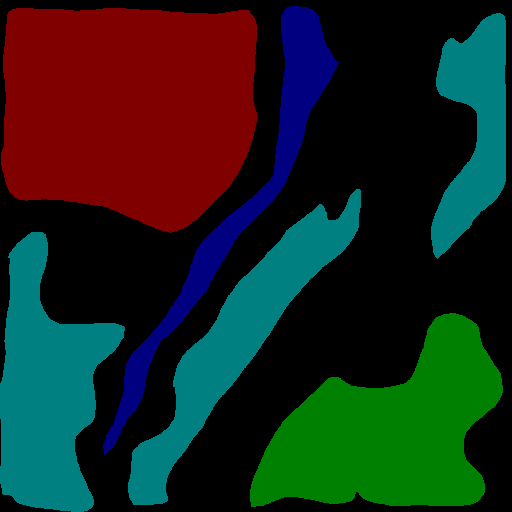}\\
            (a) LR and SG  &(b) RK and SR  &(c) BS, DS and LR &(d) SG, BS, DS and SH
        \end{tabular}
        \caption{Example SAS images with various textures: SH (blue), DS (green), BS (cyan), SG (red), RK (purple), SR (glod), LR (Gray) and unlabeled pixels (black). Textures from left to right are: (a) large sand ripple and seagrass; (b) rock and small sand ripple; (c) bright sand, dark sand and large sand ripple; (d) seagrass, bright sand, dark sand and shadow.}
        \label{fig:dset_ep}  
\end{figure*}

\begin{table}
    \normalsize
    \centering
        \begin{tabular}{ccc}
        \hline
        Class No. & Class Name      & Proportion \\ \hline
        1         & Shadow (SH)              & 0.003 \\
        2         & Dark Sand (DS)           & 0.096 \\
        3         & Bright Sand (BS)         & 0.090 \\
        4         & Seagrass (SG)             & 0.073 \\
        5         & Rock (RK)                 & 0.018 \\
        6         & Small Sand Ripple (SR)    & 0.033 \\
        7         & Large Sand Ripple (LR)  & 0.221 \\ \hline
        -         & Labeled Pixels       & 0.534 \\ 
        -         & Unlabeled Pixels     & 0.466 \\ \hline
        \end{tabular}
    \caption{The names and sample proportions of the classes in the dataset used for seabed SAS image segmentation analysis. As shown in the Table, the class distribution is highly imbalanced, especially for the Shadow class.}
    \label{tab:dataset}
\end{table}
% TODO: move the image equalization part into method.
The dataset consists of 113 high-resolution complex-valued SAS images acquired from a high-frequency (HF) synthetic aperture sonar (SAS) system deployed on an unmanned underwater vehicle (UUV). The images were originally $1001 \times 1001$ pixels in size which we downsample to $512\times512$ pixels for computational efficiency. \figurename~\ref{fig:dset_ep} shows example images from the dataset which includes a variety of different seabed textures. The dataset was semi-labled by a human operator into seven classes; Table \ref{tab:dataset} shows the class names and corresponding sample proportions. The class distribution is highly imbalanced especially for the Shadow class.

The original SAS images are single-look complex (SLC) images made up of complex-valued pixels with a large dynamic range (this is typical of the SAS modality). All the comparison methods operate on the magnitude version of the SLC and IDUS is no different.  Furthermore, each comparison method implements dynamic range compression to make the images suitable for processing of which we replicate for each respective work.  For IDUS, we dynamic range compress the magnitude image by using Schlick's \emph{rational mapping} operator \cite{schlick1995quantization}. We set the target brightness \cite{9389388} to $0.5$. Additionally, we further use the OpenCV function ``equalizeHist'' \cite{opencv_library} to equalize the gray-scale histogram of the tone-mapped images and normalize each to zero mean and unit standard deviation.

\newcommand{\blue}[1]{\textcolor{blue}{#1}}
\subsection{Convergence Analysis}\label{converge_ana}
\begin{figure}
        \centering
        \includegraphics[width=1\linewidth]{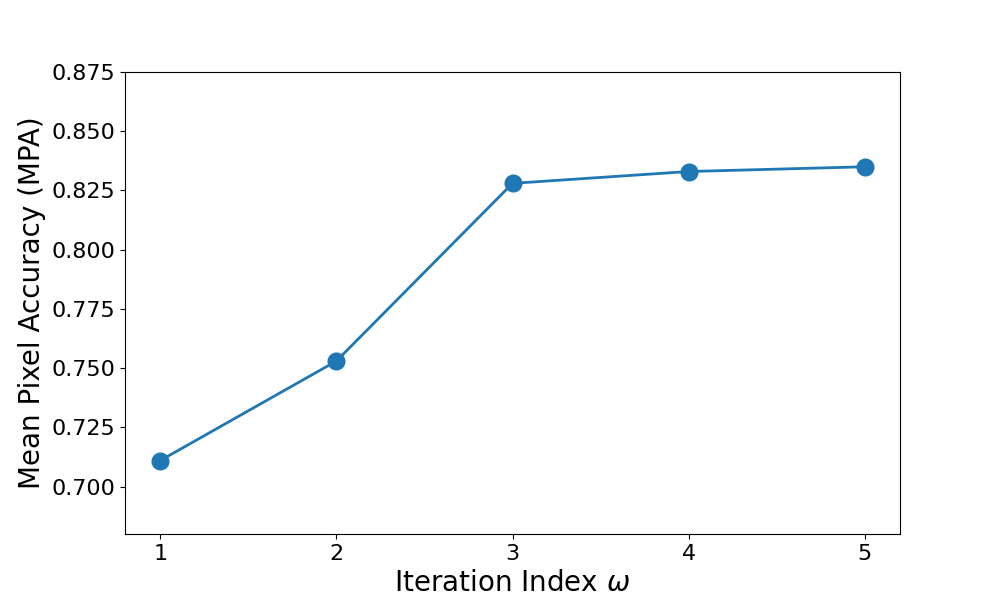}
        \caption{Mean pixel accuracy (MPA) for training images as a function of IDUS iterations (recall iteration index in Algorithm~\ref{alg:algorithm1}). As iterations progress, the MPA and hence IDUS's predicted segmentation map converges.}
        \label{fig:converge_loss}  
\end{figure}

\begin{figure*}
        \centering
        \setlength\tabcolsep{1.5pt}
        \begin{tabular}{ccccccc}
            \includegraphics[width=0.14\linewidth]{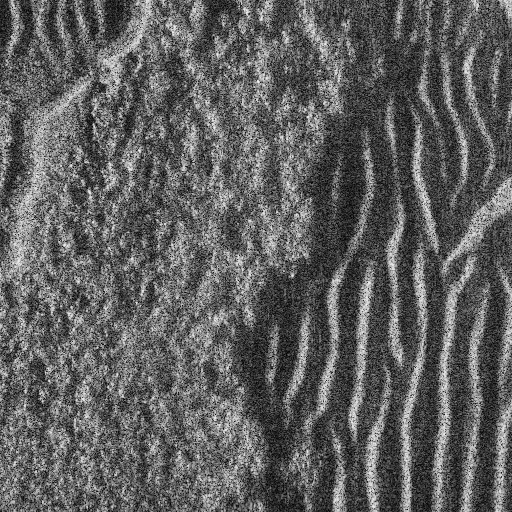}& \includegraphics[width=0.14\linewidth]{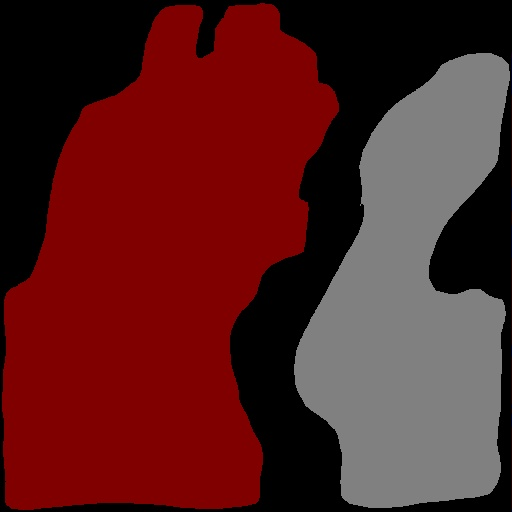}&
            \includegraphics[width=0.14\linewidth]{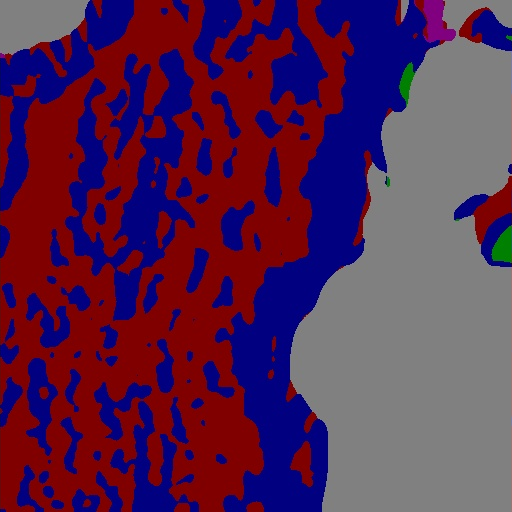}&
            \includegraphics[width=0.14\linewidth]{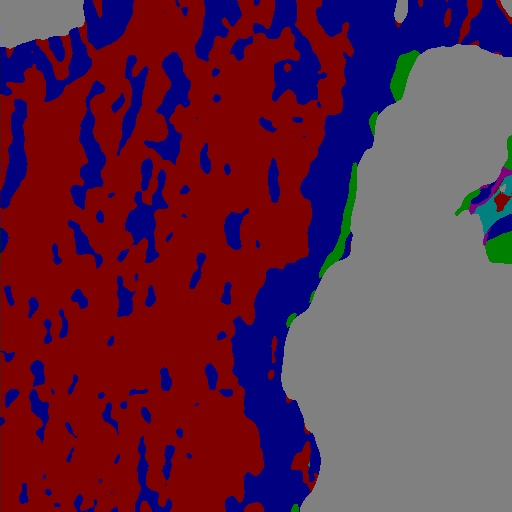}&
            \includegraphics[width=0.14\linewidth]{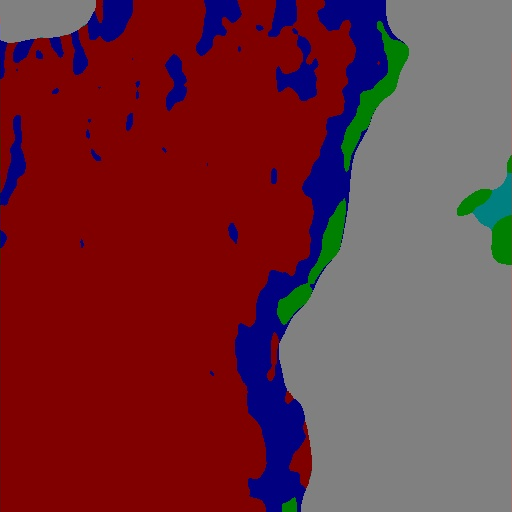}&
            \includegraphics[width=0.14\linewidth]{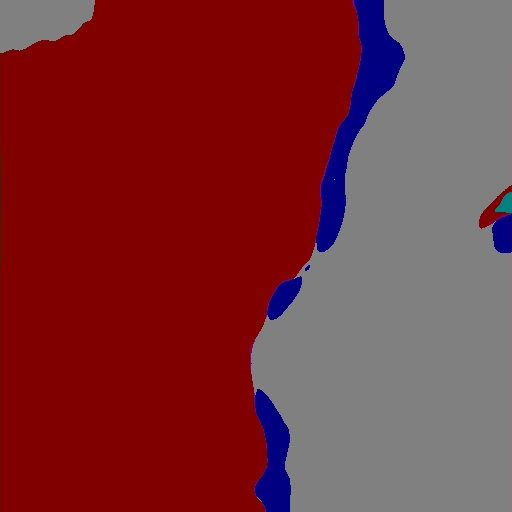}&
            \includegraphics[width=0.14\linewidth]{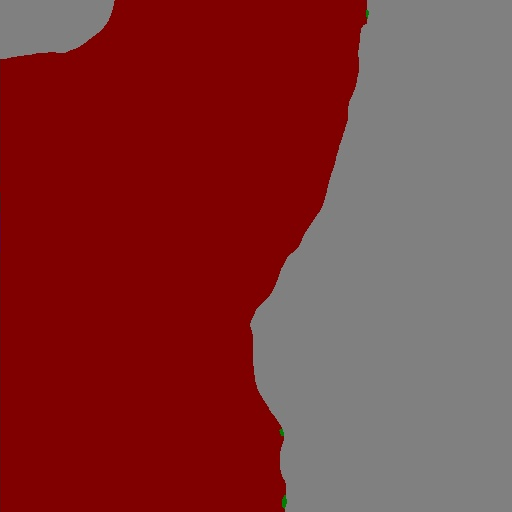}\\
            (a) Input Image  &(b) Ground Truth &(c) Iteration 1 &(d) Iteration 2 &(e) Iteration 3 &(f) Iteration 4 &(g) Iteration 5
        \end{tabular}
        \caption{The IDUS network predictions ($h_{\theta}(\mathbf{x_n})$ in \figurename~\ref{alg:algorithm1}) for an example SAS image after each iteration. Class-Color mapping is the same as \figurename~\ref{fig:dset_ep}. The results show that as the iterations proceed, the network predictions of the same textures tend to be homogeneous and finally remain unchanged and consistent with the ground truth.}
        \label{fig:converge_vis}  
\end{figure*}

To investigate the convergence of IDUS, we show the mean pixel accuracy (MPA) for training images of each iteration in \figurename~\ref{fig:converge_loss}. As shown in \figurename~\ref{fig:converge_loss}, the MPA converges as iterations progress and remains essentially unchanged for the final iteration. Moreover, in \figurename~\ref{fig:converge_vis} we visualize an example of a SAS input image, its corresponding ground truth, and the output of IDUS network for each iteration. \figurename~\ref{fig:converge_vis} demonstrates the convergence of IDUS by showing that the network predictions of similar textures tend to become homogeneous and do not have a noticeable change in the last several iterations.

\subsection{Unsupervised Co-Segmentation}\label{up_co_seg}
\begin{figure*}[t]
        \centering
        \includegraphics[width=0.85\linewidth]{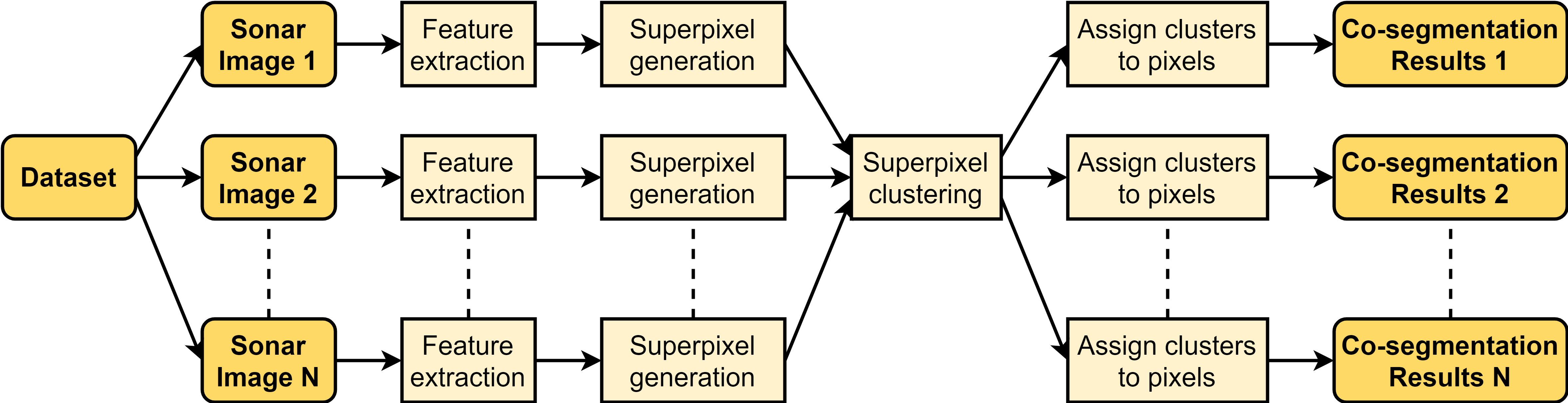}
        \caption{Block diagram of the co-segmentation strategy for hand-crafted feature descriptors. The strategy contains four main steps. \textbf{(1)} Ingest a set of feature maps from input sonar images via feature extraction. \textbf{(2)} Performing SLIC \cite{SLIC} on the extracted features to generate superpixel features by Eq \eqref{eq:1}. \textbf{(3)} Unsupervised cluster all the superpixels and provide each superpixel a cluster assignment. \textbf{(4)} Pixels inside a superpixel are assigned the same cluster by Eq \eqref{eq:2} to get pixel-level co-segmentation. }
        \label{fig:co_seg_handcrafted}  
\end{figure*}

% \textcolor{blue}{TODO- step 3 and 4 are slightly unclear).}

\begin{figure*}[t]
        \centering
        \includegraphics[width=0.85\linewidth]{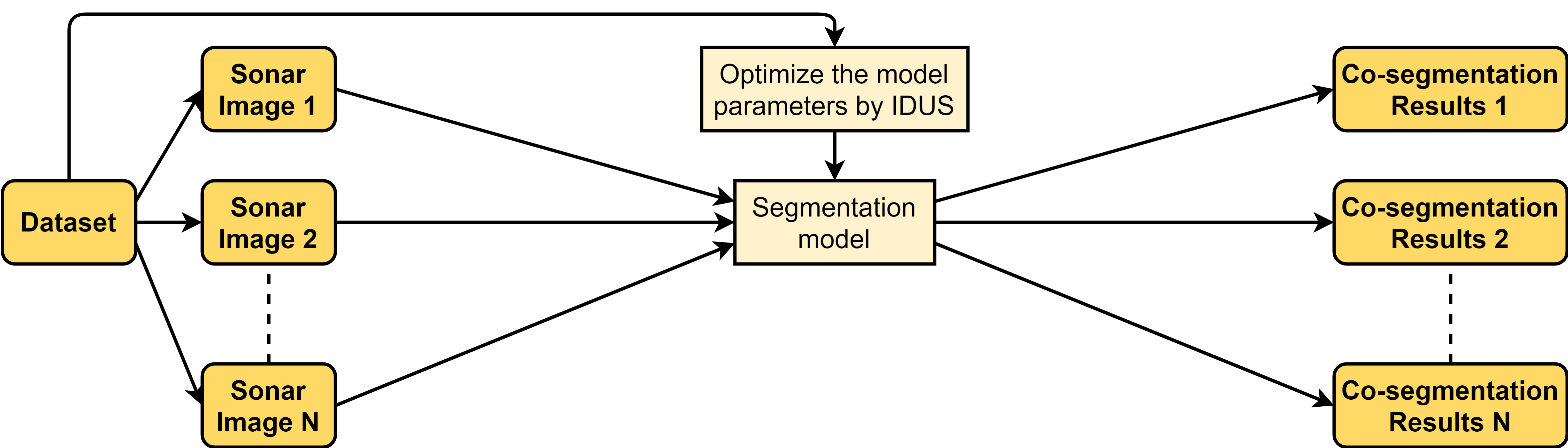}
        \caption{Block diagram of the co-segmentation strategy for IDUS. We train the IDUS on the entire dataset first, and then directly use the segmentation model outputs ingesting from the input sonar images as the final co-segmentation results. The difference between the strategy for IDUS and the strategy in \figurename~\ref{fig:co_seg_handcrafted} is that the co-segmentation of hand-crafted feature descriptor needs four steps including feature extraction, superpixel generation, superpixel clustering, and mapping superpixel cluster assignments to each pixel. However, IDUS integrates these four steps into a single model which is optimized.}
        \label{fig:co_seg_idus}  
\end{figure*}

\begin{figure*}
    \centering
    \begin{tabular}{cc}
     \includegraphics[width=0.55\linewidth]{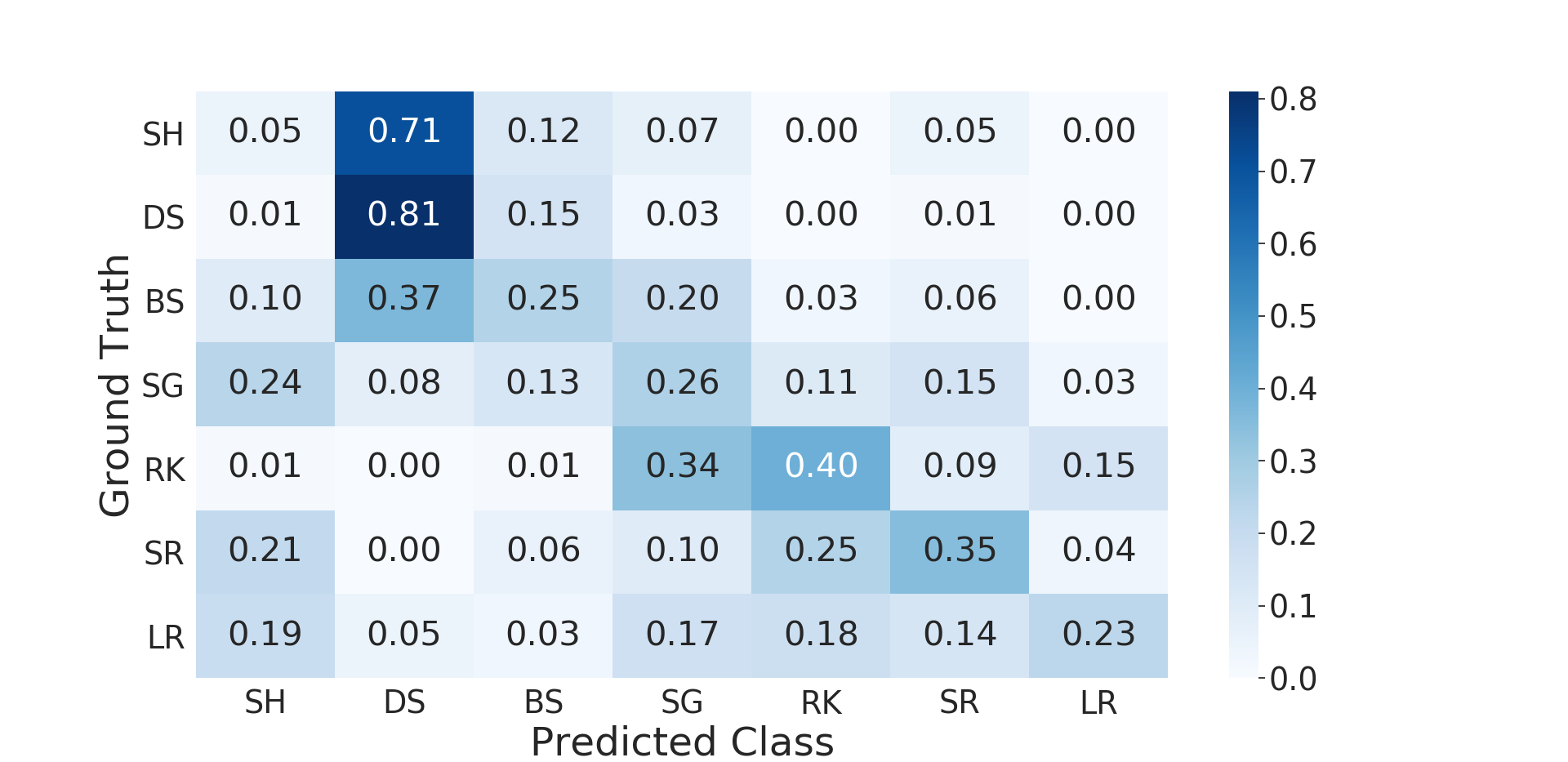} & \includegraphics[width=0.55\linewidth]{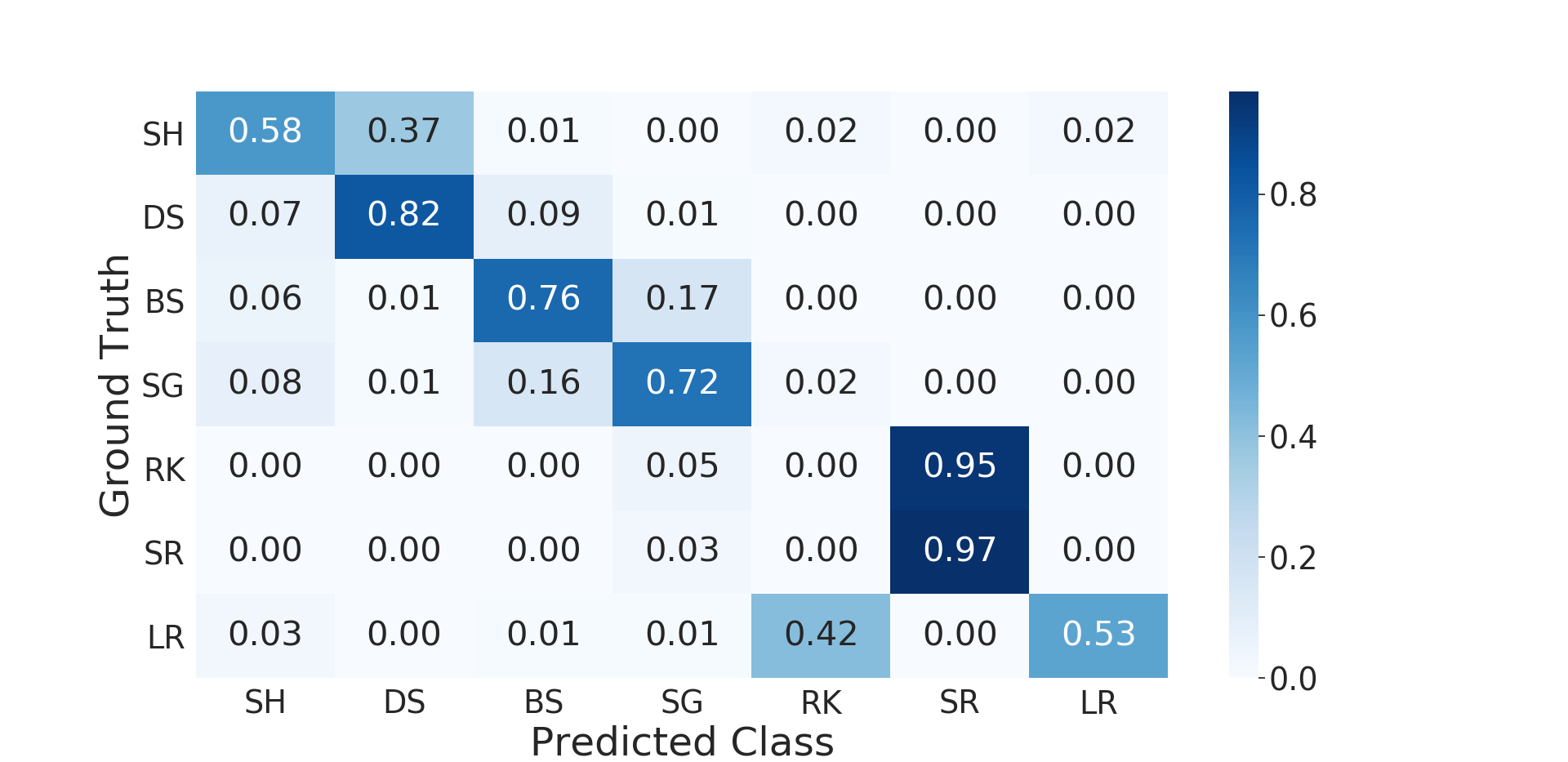}\\
        (a) Lianantonakis, \textit{et al.} (2007) \cite{lianantonakis2007sidescan} & (b) Williams (2009) \cite{williams2009unsupervised} \\
        \includegraphics[width=0.55\linewidth]{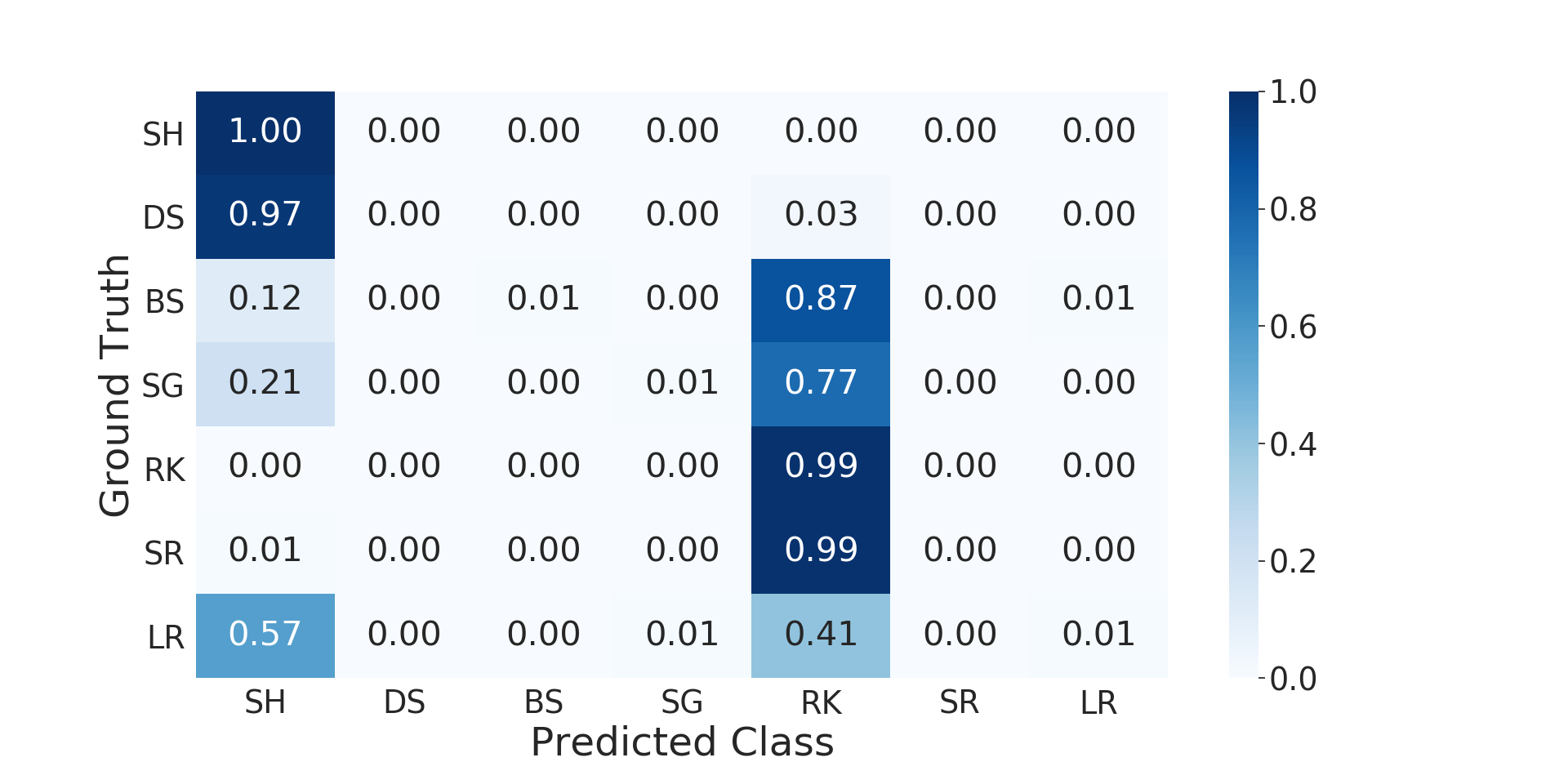} & \includegraphics[width=0.55\linewidth]{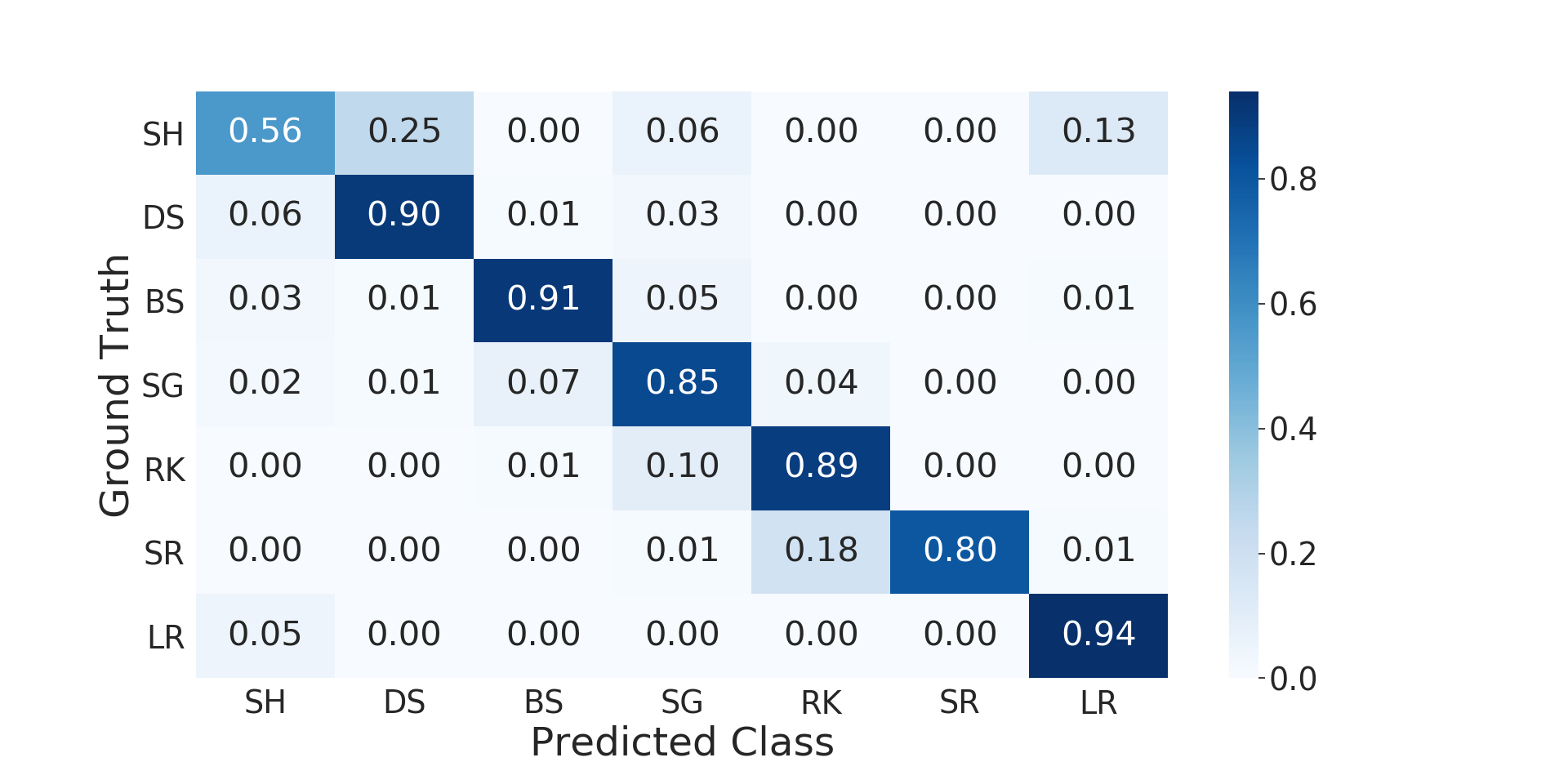} \\
        (c) Zare, \textit{et al.} (2017) \cite{Zare2017PossibilisticFL} &  (d) IDUS\\
       
    \end{tabular}
    \caption{Confusion matrices of IDUS along with comparisons against state of the art unsupervised methods.}
    \label{fig:confusion_matrix}  
\end{figure*}

In image processing, the term co-segmentation is a special case of image segmentation which is defined as a joint pixel-level segmentation of semantically similar objects over a set of given images \cite{5995530}. The objects can be a set of small rectangular image patches \cite{williams2009unsupervised}, or superpixels aligning the image textures' boundaries well \cite{Zare2017PossibilisticFL}. In SAS image segmentation \cite{williams2009unsupervised, Zare2017PossibilisticFL}, co-segmentation is a popular way to differentiate seabed environment textures mainly due to the difficulty of obtaining labeled data since co-segmentation does not rely on labeled data. Therefore to show the effectiveness of IDUS, we compare performance with several comparison methods based on two co-segmentation strategies, one common to the comparison methods and one we tailored for IDUS. Since data labels are not used for co-segmentation, the entire dataset (113 sonar images) shown in \tablename~\ref{tab:dataset} is used in the analysis. 

\noindent \textbf{Comparison Methods}: We compare with three state-of-the-art methods \cite{lianantonakis2007sidescan, williams2009unsupervised,Zare2017PossibilisticFL} used in SAS seabed environment image segmentation. Following, we give the implementation details we use in generating the comparison methods as no source code is publicly available to evaluate. We make a best effort attempt to reproduce the methods as given in their respective sources. Our implementation is based on Python $3.7$ code running on an Intel (R) Core (TM) i9-7960X 2.80GHz CPU with Linux operating system.

\textbf{Lianantonakis, \textit{et al.} (2007) \cite{lianantonakis2007sidescan}.} This method uses Haralick features \cite{haralick1973textural} derived from the gray-level co-occurrence matrix (GLCM) and couples this with active contours to arrive at a binary class mapping. 
We extend this work to multiple classes by simply using the same feature descriptors as the original work but apply $k$-means++ \cite{arthur2006k} to cluster; a similar replication approach is used in \cite{cobb2011autocorrelation}. We ran $k$-means++ with 100 random initializations and selected the run that produces the minimum within cluster sum of squares error in a manner consistent with \cite{williams2009unsupervised}. 

\textbf{Williams (2009) \cite{williams2009unsupervised}.} This method uses wavelet features along with spectral clustering to compute the segmentation map.  We found that spectral clustering results in similar performance as using $k$-means++ so we opt to use for simplicity as we did in Lianantonakis, \textit{et al.} mentioned above; a similar replication approach is used in \cite{cobb2011autocorrelation}.

\textbf{Zare, \textit{et al.} (2017) \cite{Zare2017PossibilisticFL}.} In this work, the feature sets are produced by Sobel edge descriptors (Sobel) \cite{4610973}, histograms of oriented gradients (HOG) \cite{dalal2005histograms}, and local binary pattern (LBP) features \cite{guo2010completed}. For each feature descriptor, we use the same sliding window strategy of Lianantonakis, \etal \cite{lianantonakis2007sidescan} to derive a feature vector for each pixel. 

For the comparison methods which use hand-crafted feature descriptors, we summarize the co-segmentation strategy in \figurename~\ref{fig:co_seg_handcrafted} including four main steps:
\begin{enumerate}
\item Feature extraction to extract a set of feature maps from the input sonar images. 
\item Performing SLIC \cite{SLIC} on the extracted features to generate superpixel features by Eq \eqref{eq:1}. 
\item Unsupervised cluster all the superpixels and provide each superpixel a cluster assignment. 
\item Pixels inside a superpixel are assigned the same cluster by Eq \eqref{eq:2} to get pixel-level co-segmentation.
\end{enumerate}
 
% need to textit argmax?
For the co-segmentation strategy of IDUS shown in \figurename~\ref{fig:co_seg_idus}, we train a segmentation network by IDUS on the entire dataset first, and then directly use the network outputs computed from the input sonar images as the final co-segmentation results. %Since IDUS is unsupervised, in the scenario of the co-segmentation, it is fair to use the prediction of training images as the co-segmentation results. 
The difference between the strategy for IDUS (\figurename~\ref{fig:co_seg_idus}) and the strategy in \figurename~\ref{fig:co_seg_handcrafted} is that the co-segmentation of the comparison methods (i.e. hand-crafted feature descriptors) requires four steps including feature extraction, superpixel generation, superpixel clustering, and mapping superpixel cluster assignments to each pixel. However, IDUS integrates these four steps into one single model (i.e. learning algorithm) and is performed jointly while the model is optimized.

We use the confusion matrix as a criterion to evaluate the co-segmentation performance of IDUS and the comparison methods on the entire dataset (113 images). A confusion matrix shows the proportion of an assigned class predicted by the algorithm for a given ground truth class. Ideally, the ground truth class and the predicted class overlap entirely giving a proportion of one. However, in practice, the predicted class by co-segmentation usually do not perfectly match with the ground truth class resulting in a proportion less than one. 

As a result of performing unsupervised segmentation, class assignments by the algorithm must be mapped to ground-truth classes. We perform this assignment by initially using a random disjoint assignment among the possible classes and then compute the confusion matrix.  Next, we re-sort the columns of the matrix in such a way so that the sum of diagonal elements in the new confusion matrix is maximized. For the co-segmentation of the comparison methods, we set the clusters number is the same as the number of ground-truth classes (seven class). For IDUS, we set the dimension of the network softmax output as seven. Thus, we use a confusion matrix with size $7\times7$ as the criterion for evaluating different methods. Since the ground truth is semi-labeled, the unlabeled pixels do not contribute to the procedure of computing confusion matrix during testing (recall the proportion of unlabeled pixels in \tablename~\ref{tab:dataset}).

\figurename~\ref{fig:confusion_matrix} shows the confusion matrices of IDUS and the comparison methods. \tablename~\ref{tab:cfm_mean} reports each method's mean pixel accuracy (MPA), which is derived by the mean of diagonal elements in the confusion matrix. As shown in the \figurename~\ref{fig:confusion_matrix} and \tablename~\ref{tab:cfm_mean}, IDUS results in superior co-segmentation results (\figurename~\ref{fig:confusion_matrix}d) over the comparison methods (\figurename~\ref{fig:confusion_matrix}a,b and c).

Upon examining the confusion matrix, we notice some comparison methods could not provide enough discrimination ability to differentiate certain classes well. For example, Rock (Rk) and Small Sand Ripple (SR) are clustered into the same class in Williams \cite{williams2009unsupervised} (\figurename~\ref{fig:confusion_matrix}b). However, such misclassification problems are diminished by IDUS (\figurename~\ref{fig:confusion_matrix}d).
\begin{table}
    \normalsize
    \centering
        \begin{tabular}{cc}
        \hline
        Method        & MPA \\ \hline
        Haralick      & 0.336 \\
        Wavelet       & 0.626 \\
        SHL           & 0.288 \\
        \textbf{IDUS}          & \textbf{0.836} \\ \hline
        \end{tabular}
    \caption{Mean pixel accuracy (MPA) of IDUS along with comparisons in the experiment of \figurename~\ref{fig:confusion_matrix}. The MPA is derived by the mean of diagonal elements in the confusion matrix. A higher value indicates better performance.}
    \label{tab:cfm_mean}
\end{table}

\begin{figure*}
    \centering
    \includegraphics[width=6.5in]{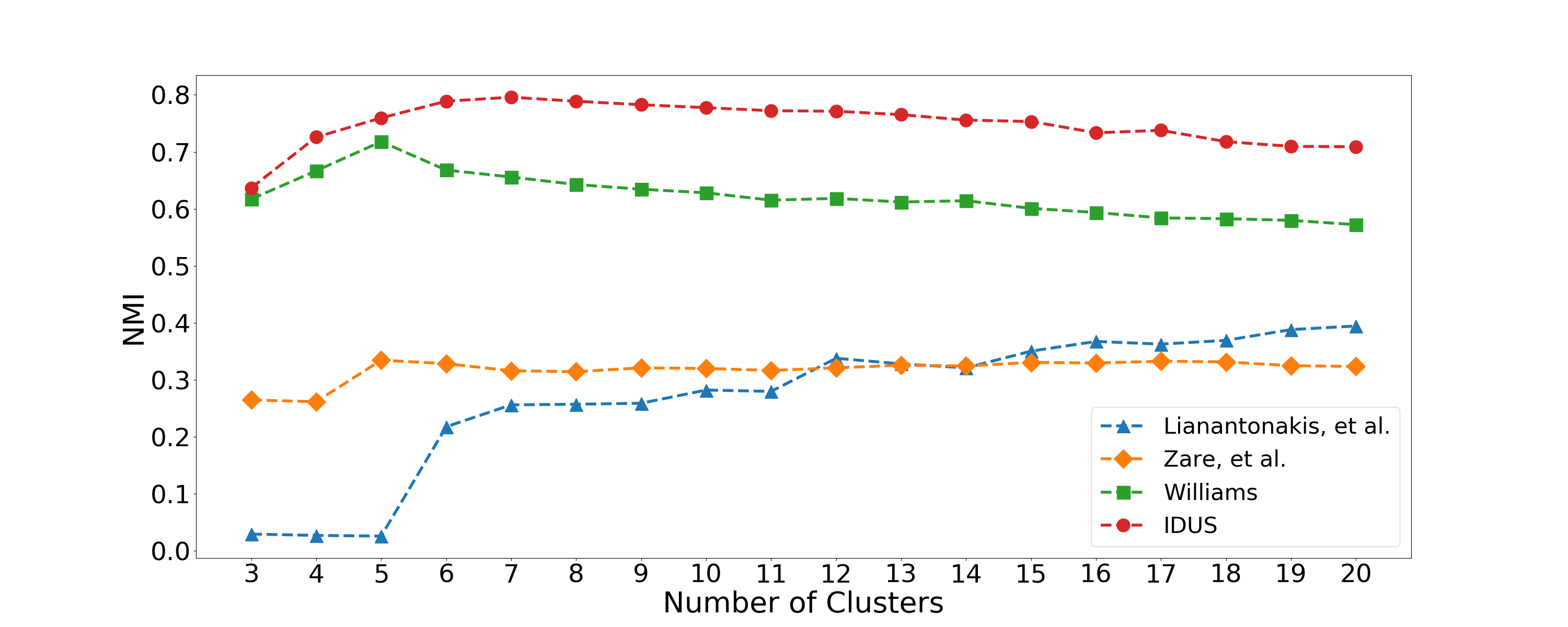}
    \caption{Normalized mutual information (NMI) for various choices of cluster numbers. The results demonstrate that IDUS performs better than competing alternatives, independent of the number of clusters.} 
    \label{fig:nmi}
\end{figure*}

Although IDUS demonstrates the best performance in confusion matrices, the evaluation assumes each class only contains one texture type. In fact, some features might be multi-modal. In other words, one class may contain several feature types resulting in a a single class being distributed among several subclasses in the latent space. For example, the category of Large Sand Ripple (LR) contains two different scales of ripple textures (in \figurename \ref{fig:dset_ep}a and b, the scales of LRs appear different). Besides this, the similarities between texture types are also different and some classes can be grouped into more general classes (e.g. Bright Sand and Dark Sand can be grouped into "Sand", and Large Sand Ripple and Small Sand Ripple can be grouped into "Sand Ripple"). Therefore, the number of classes, (recall we use seven classes/clusters) may be sub-optimal for grouping textures in feature space, and it is necessary to evaluate the performance of the co-segmentation with different numbers of clusters. However, due to the limitations of the confusion matrix, the evaluation results are not easy to analyze when the co-segmentation number of clusters are different from the number of ground truth classes. To address this, we use normalized mutual information \cite{caron2018deep} as a new metric to replace the confusion matrix for evaluating the co-segmentation results with various clusters numbers.

Normalized mutual information (NMI) is a metric used in DeepCluster \cite{caron2018deep} to evaluate the performance of unsupervised clustering methods. From the view of information theory, the NMI measures the information shared between the predicted classes and the ground truth classes for unsupervised segmentation. A higher NMI indicates better performance since the information overlap between the predicted classes and the ground truth classes is higher. NMI can be directly derived from a contingency matrix without the consideration of class assignment mismatch problem existing in \figurename~\ref{fig:confusion_matrix}. In other words, we can use a contingency matrix without column re-sorting to compute NMI. 

More precisely, given a set of unsupervised predicted classes, $A$, and ground truth classes, $B$, the NMI is defined as $\mathit{NMI}(A;B) = \dfrac{\mathit{I}(A;B)}{\sqrt{\mathit{H}(A)\mathit{H}(B)}}$,
where $\mathit{I}(A;B)$ is the mutual information between $A$ and $B$, $\mathit{H(\cdot)}$ is a measure of the entropy of classes distribution.

To compute NMI, we extract the decoder outputs from the segmentation network (the model used for the evaluation in \figurename~\ref{fig:confusion_matrix}d) of IDUS as a new set of feature maps after training and use the same co-segmentation strategy shown in \figurename~\ref{fig:co_seg_handcrafted}. This results in different co-segmentation results each corresponding to different number of unsupervised predicted classes.  
We show co-segmentation of IDUS and comparison methods using $k$-means clustering of the learned feature space (recall \figurename~\ref{fig:co_seg_handcrafted}, it is superpixel clustering) over a range of the number of clusters from three to twenty. This range of cluster cardinality allows us to explore the texture types from coarse to fine. Recall the implementation details of Zare, \etal \cite{Zare2017PossibilisticFL}, in which they use PFLICM for unsupervised clustering. We found PFLICM does not match the dataset used in our work (see  \figurename~\ref{fig:confusion_matrix}c). For a fair comparison, we use $k$-means instead of PFLICM \cite{Zare2017PossibilisticFL} to cluster the superpixels of this method.

\figurename~\ref{fig:nmi} shows the NMI scores of each method for a given number of clusters. The results demonstrate that IDUS performs better than the comparison methods regardless of the number of clusters. In other words, IDUS features have a more powerful feature representation for seabed environment textures. We notice that the NMIs of the Lianantonakis, \etal \cite{lianantonakis2007sidescan} are lower in the case of small cluster numbers and higher in large cluster numbers. Such a phenomenon demonstrates that the features from this method are multi-modal for some types of textures requiring several sub-clusters to be grouped to yield good results for the final assigned classes.

\subsection{Feature and Training Robustness}
\label{feature_quality}

Although the co-segmentation strategy is widely used in SAS image segmentation \cite{williams2009unsupervised,Zare2017PossibilisticFL}, it is not convenient to process  new sonar images outside the training dataset: the strategy must add the new images into the existing dataset and perform the whole process again to obtain new co-segmentation results. To mitigate this issue, we can label some of the sonar images and train a linear classifier on the extracted features which are trained on the initial dataset exclusively.
In this manner, as new images arrived, they are segmented using this classifier.  If the new images are labeled, the existing feature extractor can be utilized (which is computational expensive to regenerate) but a new linear classifier can be computed (which is computational inexpensive to regenerate).

Inspired by the experiment design of \cite{zhang2017split}, we use labeled data to train a linear classifier on top of frozen feature sets to evaluate the feature quality or discriminative ability of the learned unsupervised features. If the features provide correct discrimination among ground truth classes, the linear classification is expected to yield good results.  Here, our task is a pixel-level classification so we directly train a linear classifier based on every pixel feature.  This is a modification from  \cite{zhang2017split} which trains a linear classifier on the feature maps of images.

Our linear classifier is created by using $1\times1\times D$ convolutional kernels to implement $C$ linear functions, where $C$ is the number of classes and $D$ the feature dimensions. We separate the entire data set into a train-test split of $7:3$ giving eighty images for the training set and thirty-three for the testing set. The class proportions for training set and testing set are shown in Table \ref{tab:train_set} and \ref{tab:test_set}, respectively. We first train IDUS in an unsupervised fashion using only the training set.  Next,  we use the training set ground truth to train a linear classifier on the decoder outputs of the frozen IDUS network. For the comparison methods (Section \ref{up_co_seg}), we use the feature extraction step of the co-segmentation strategy shown in the \figurename~\ref{fig:co_seg_handcrafted} to extract a set of feature maps from the training set for each method. Next, we use ground truth to train a linear classifier on the feature maps using the same classifier specifid above. The loss function and optimizer are the same as the training configurations of Section \ref{train_config}; however, the learning rate is started from $0.01$ and decreased by $0.1$ every thirty epochs. Since the ground truth is semi-labeled, the unlabeled pixels are not be used to compute the loss function (i.e. their associated loss has a weight of zero). During the evaluation of the testing set, we extract the feature from the testing sonar images (for IDUS, the feature is the decoder outputs of frozen segmentation network; for the comparison methods, the feature extraction uses a hand-crafted feature via a sliding window to extract feature) and then input them to the classifier to get the final segmentation result.

\begin{table}
    \normalsize
    \centering
        \begin{tabular}{ccc}
        \hline
        Class No. & Class Name      & Proportion \\ \hline
        1         & Shadow (SH)              & 0.002 \\
        2         & Dark Sand (DS)           & 0.098 \\
        3         & Bright Sand (BS)         & 0.097 \\
        4         & Seagrass (SG)             & 0.078 \\
        5         & Rock (RK)                 & 0.018 \\
        6         & Small Sand Ripple (SR)    & 0.038 \\
        7         & Large Sand Ripple (LR)  & 0.215 \\ \hline
        -         & Labeled Pixels       & 0.546 \\ 
        -         & Unlabeled Pixels     & 0.454 \\ \hline
        \end{tabular}
    \caption{The names and sample proportions of different classes in the training set (eighty sonar images) used for training the classifiers of unsupervised features.}
    \label{tab:train_set}
\end{table}
\begin{table}
    \normalsize
    \centering
        \begin{tabular}{ccc}
        \hline
        Class No. & Class Name      & Proportion \\ \hline
        1         & Shadow (SH)              & 0.003 \\
        2         & Dark Sand (DS)           & 0.091 \\
        3         & Bright Sand (BS)         & 0.072 \\
        4         & Seagrass (SG)             & 0.063 \\
        5         & Rock (RK)                 & 0.019 \\
        6         & Small Sand Ripple (SR)    & 0.023 \\
        7         & Large Sand Ripple (LR)  & 0.236 \\ \hline
        -         & Labeled Pixels       & 0.507 \\ 
        -         & Unlabeled Pixels     & 0.493 \\ \hline
        \end{tabular}
    \caption{The names and sample proportions of different classes in the testing set (thirty-three sonar images) used for evaluating the performance of unsupervised feature classifiers.}
    \label{tab:test_set}
\end{table}

We use pixel accuracy (PA) for each class and the corresponding mean pixel accuracy (MPA) to evaluate the performance of the linear classifiers. The equation to compute the pixel accuracy $P_c$ for class $c$ is defined by $P_c = \dfrac{\sum_{i=1}^{N_p}I(a_i==c,~b_i==c)}{\sum_{i=1}^{Np}I(b_i==c)}$,
where $a_i$ and $b_i$ is the classifier prediction and ground truth of $i$th pixel, respectively, and $N_p$ is the total number of pixels. The mean pixel accuracy (MPA) can be computed by $MPA = \dfrac{1}{C}\sum_{c=1}^{C}P_c$,
where $C$ is the total number of classes. 

Table \ref{tab:linear} shows the PAs and MPA for each method on the testing set shown in \tablename~\ref{tab:test_set}. Since the dataset is semi-labeled, unlabeled pixels are not considered in the evaluation. As shown in the table, IDUS has the best feature quality over all the classes except for the Shadow class. Note that Zare, \etal \cite{Zare2017PossibilisticFL} has an accuracy of $1.0$ for the Shadow class which is a result of assigning the majority of pixels to this class.

\begin{table*}
    \centering
    \normalsize

    \begin{tabular}{@{}ccccccccc@{}}
    \toprule
    \multirow{2}{*}{Method} & \multicolumn{7}{c}{Pixel Accuracy} &  \multirow{2}{*}{MPA} \\ \cmidrule(lr){2-8} 
        & SH   & DS   & BS   & SG   & RK   & SR   & LR  \\ \midrule
    Lianantonakis, \textit{et al.} \cite{lianantonakis2007sidescan}       & 0.00 & 0.64 & 0.46 & 0.17 & 0.67 & 0.69 & 0.79 & 0.49 \\
    Williams \cite{williams2009unsupervised}        & 0.72 & 0.04 & 0.84 & 0.23 & 0.85 & 0.25 & 0.97 & 0.56 \\
    Zare, \textit{et al.} \cite{Zare2017PossibilisticFL}     & \textbf{1.00} & 0.00 & 0.21 & 0.54 & 0.62 & 0.00 & 0.01 & 0.34 \\
    IDUS  & 0.81 & \textbf{0.87} & \textbf{0.87} & \textbf{0.69} & \textbf{0.96} & \textbf{0.45} & \textbf{0.98} & \textbf{0.80} \\ \bottomrule
    \end{tabular}
\caption{The pixel accuracies (PAs) of each class along with the corresponding mean pixel accuracy (MPA) for each method evaluated on the testing set (thirty-three sonar images, see in table \ref{tab:test_set}). Since the dataset is semi-labeled, unlabeled pixels are not evaluated. A higher value indicates better performance. Best results for each class and best MPA are in bold. IDUS has the highest PAs for most classes and the best MPA demonstrating its superior performance over competing methods.}
\label{tab:linear}
\end{table*}

\begin{figure}[t]
    \centering
    \includegraphics[width=1\linewidth]{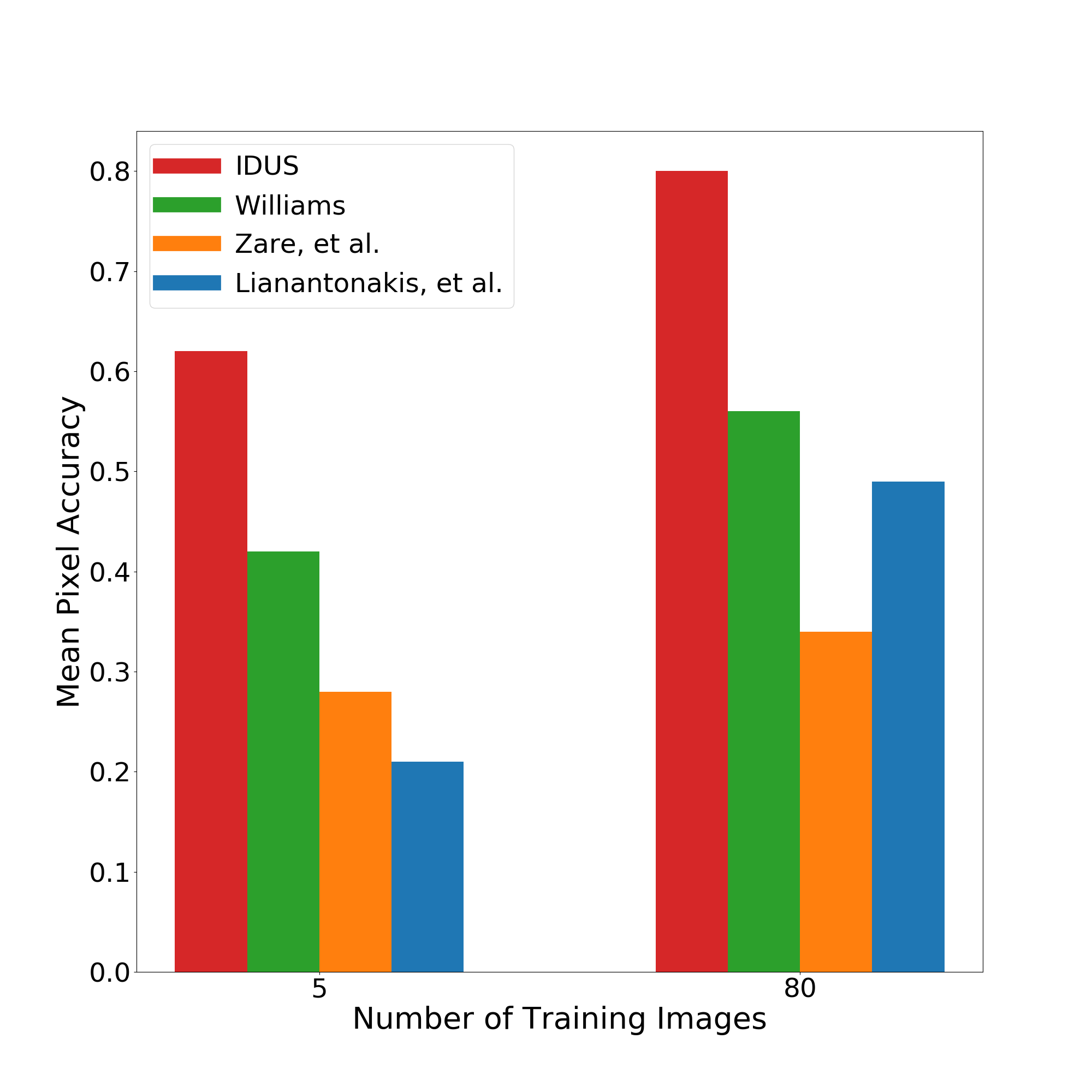}
    \caption{To better understand the performance in low training data scenarios, we train each method using the whole training set and then train using a random $6.25\%$ subset. The train-test split is with a ratio of $7:3$ giving eighty images for the training set (\tablename~\ref{tab:train_set}) and thirty-three for the testing set  (\tablename~\ref{tab:test_set}). The training subset is five images randomly selected from the training set. We report the testing set MPAs for each method trained on the training subset (five images) and the whole training set (eighty images). Higher numbers indicate better performance. As shown in the figure, the benefits of IDUS over state of the art are even more pronounced when training is limited.}
    \label{fig:low_training}
\end{figure}

\begin{figure}[t]
    \centering
    \includegraphics[width=1\linewidth]{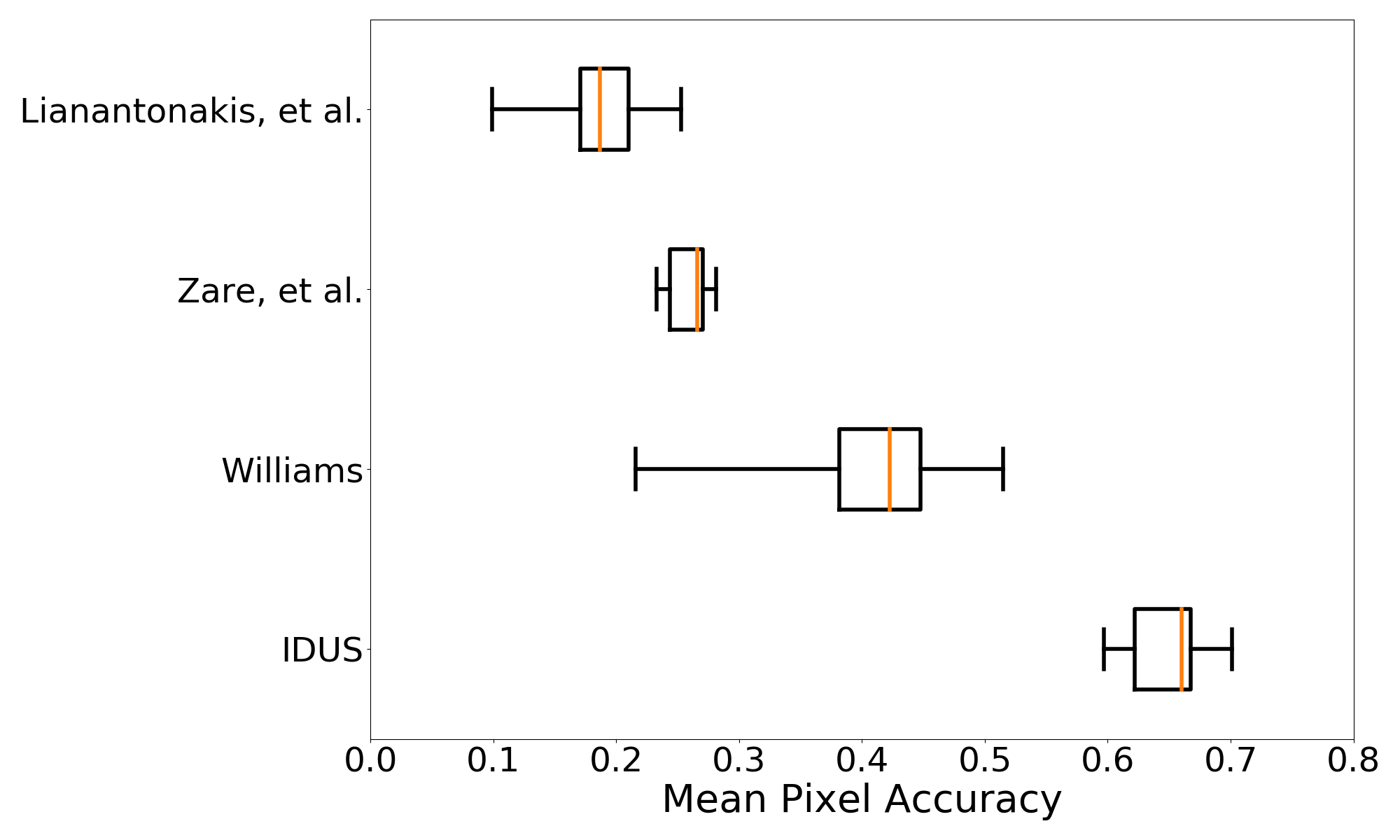}
    \caption{To evaluate the effects of training set selection bias, we use the same experimental setup for results shown in \figurename~\ref{fig:low_training}, but conduct each experiment ten times using random subsets of $6.25\%$ (five images) of the eighty training images. As shown, IDUS presents the best mean pixel accuracy of all the methods; higher numbers indicate better performance.}
    \label{fig:sel_bias}
\end{figure}

\noindent \textbf{Performance Vs. Training Size:} Deep learning-based algorithms usually work well when abundant training data is provided. However, when little labeled data is available, the performance may dramatically decline due to over-fitting. Therefore, to better understand the performance of IDUS and competing state of the art methods in low training data scenarios, we randomly select five images from the training set (\tablename~\ref{tab:train_set}) and extract the corresponding feature maps for each method. Next, we train a linear classifier on the extracted five feature maps for each method and test the linear classifier on the testing set (\tablename~\ref{tab:test_set}). \figurename~\ref{fig:low_training} shows the testing set MPAs of training on five and eighty images. As shown in the figure, the large MPA gap between IDUS and the second-best method Williams \cite{williams2009unsupervised} indicates IDUS achieves superior results even in the low training data scenario.

Although we show IDUS obtains good performance in the low training data scenario, the effectiveness may be from selection bias of the training samples. To examine this issue, we repeat the same experiment above ten times using a random subset of $6.25\%$ (five images) of the eighty training images. We report the results as MPA in \figurename~\ref{fig:sel_bias}. It is noteworthy to mention that we use the same training subsets for the evaluation of each algorithm. As revealed by \figurename~\ref{fig:sel_bias}, IDUS achieves the highest mean pixel accuracy with little variation based on the exact choice of training image set.

\subsection{Computational Burden}
\label{compute_pfm}

\begin{table}
    \normalsize
    \centering
        \begin{tabular}{cc}
        \hline
        % idg - putting these to more accurate sig figs
        Method        & Run-Time [s/image]\\ \hline
        Lianantonakis, \etal \cite{lianantonakis2007sidescan}      & $858$ \\
        Williams \cite{williams2009unsupervised}       & $31$ \\
        Zare, \etal \cite{Zare2017PossibilisticFL}           & $4.6$ \\
        IDUS (CPU)    & $1.0\times10^{-1}$ \\
        IDUS (GPU)    & $7.0\times10^{-3}$ \\\hline
        \end{tabular}
    \caption{The inference time of each method in seconds per image during testing; lower numbers indicate better performance. The run-time includes the time of feature extraction and linear classification (recall in Section \ref{feature_quality}). To provide a fair evaluation against the comparison methods (which only run on a CPU), we report the running time of IDUS on a GPU and CPU. As shown, IDUS is faster than the comparison methods by at least an order-of-magnitude.}
    \label{tab:compute_time}
\end{table}

We evaluate the run-time of all the methods and display the results in \tablename~\ref{tab:compute_time}. The run-time includes the time of feature extraction and linear classification (recall Section \ref{feature_quality}). The table reports the average inference time of each method in seconds per image of the test set; lower numbers indicate better performance. Since the comparison methods (Lianantonakis, \etal \cite{lianantonakis2007sidescan}, Williams \cite{williams2009unsupervised}, and Zare, \etal \cite{Zare2017PossibilisticFL}) are designed for execution on a CPU architecture, we also report the inference time of IDUS on the CPU to provide a fair comparison. As shown in \tablename~\ref{tab:compute_time}, the inference speed of IDUS is much faster (even in the worse case, on CPU) than the second-best method Zare, \etal \cite{Zare2017PossibilisticFL}. The fastest speed is obtained by running IDUS on a GPU and is sufficient for real-time application.
\subsection{Comparisons Against Supervised Deep Learning Methods}
\label{idssevl}

\begin{figure*}[t]
    \centering
    \includegraphics[width=6.0in]{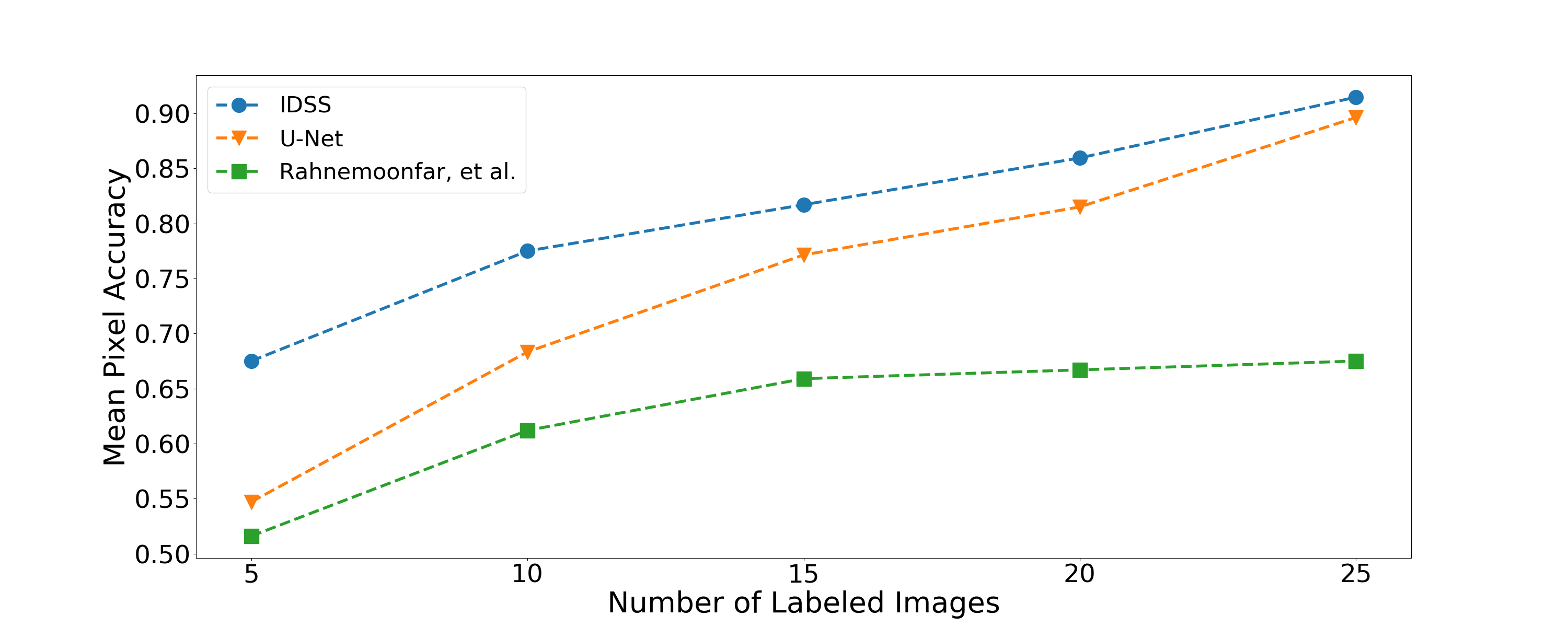}
    \caption{The mean pixel accuracy (MPA) for different training set sizes; higher values indicate better performance. IDSS outperforms comparison methods, especially in the case of very limited training (five samples).}
    \label{fig:exp4}
\end{figure*}

Following Algorithm \ref{alg:comb}, we use IDSS to fine-tune the network on using different size training sets of labeled data set to explore the performance improvement of combining IDUS with supervised training. The train-test splits we used are the same as Section \ref{feature_quality}. In the unsupervised learning step (Step 2 in Algorithm 2), we use eighty sonar images (whole training set) without ground truths $\{X_{unlabel}\}$ to learn features by IDUS. In the supervised learning step (Step 4 in Algorithm 2), we randomly pick a subset of labeled images $\{X_{label}, Y_{label}\}$ from the training set to fine-tune the U-Net.

\figurename~\ref{fig:exp4} shows the plot of MPAs on different training set sizes of labeled images $\{5, 10, 15, 20, 25\}$. We also train a state of the art U-Net \cite{unet} and the recent supervised approach of Rahnemoonfar \textit{et al.} \cite{rahnemoonfar2019semantic} on different training set sizes to compare the performance of best-known supervised deep networks to IDSS \footnote{The IDSS network and the supervised U-Net are architecturally identical but of course with distinct optimized parameters. We also design Rahnemoonfar \textit{et al.} \cite{rahnemoonfar2019semantic} consistent with the description in their paper and to have roughly the same number of parameters as the supervised U-Net.}. For fairness in comparison, when training U-Net, Rahnemoonfar, \textit{et al.} and IDSS,  the training image sets, numerical optimizer, learning rate strategy and batch size are all the same as the training configurations shown in Section \ref{train_config}. 
To be clear, U-Net and Rahnemoonfar, \textit{et al.} are trained with known ground truth pixel labels. However, since a given training image is only partially-labeled, the unlabeled pixels (recall in \tablename~\ref{tab:train_set}) are not considered in the loss function computation when training the comparison models.

We do not report the accuracy of training set sizes larger than twenty-five since the accuracy does not improve much beyond this size. \figurename~1\ref{fig:exp4} results once again demonstrates that IDSS benefits are even more pronounced when training data is not abundantly available. Note especially that in the lowest training data scenario of five images, IDSS provides the largest improvement (0.128 MPA) over the second best method. Note also that Rahnemoonfar, \textit{et al.} \cite{rahnemoonfar2019semantic} does not perform particularly well on our SAS dataset. We conjecture this for a couple of reasons: 1.) their approach was designed for sidescan sonar and 2.) being a purely supervised method, the method in \cite{rahnemoonfar2019semantic} is highly reliant on generous training, which is practically infeasible and not available in our SAS dataset.
\begin{figure*}
        \centering
        \setlength\tabcolsep{1.5pt}
        \begin{tabular}{ccccc}
            \includegraphics[width=0.2\linewidth]{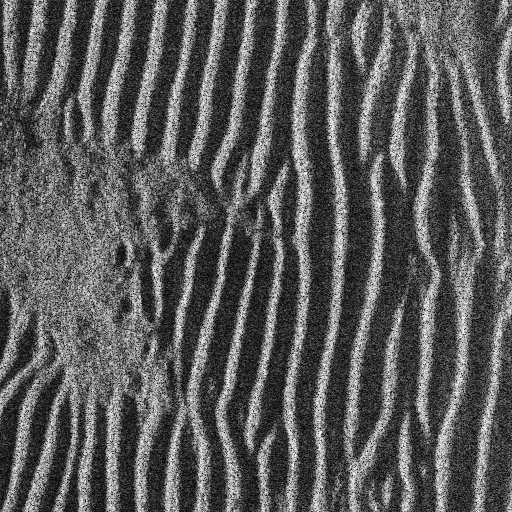}& \includegraphics[width=0.2\linewidth]{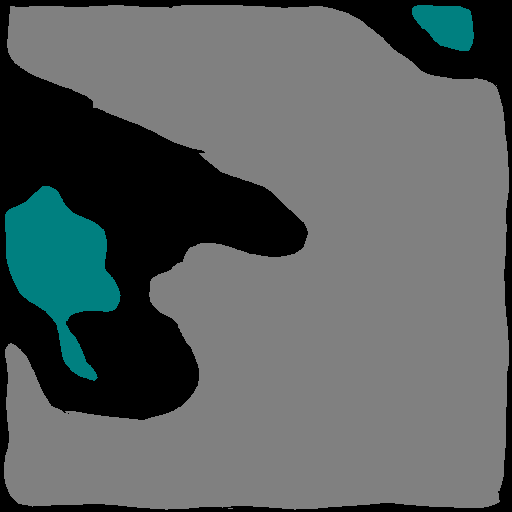}&
            \includegraphics[width=0.2\linewidth]{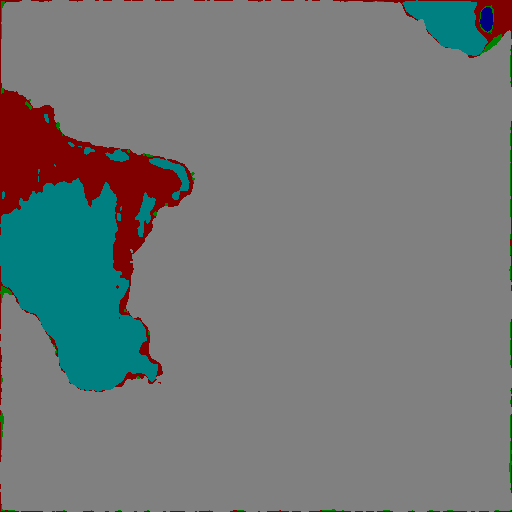}&
            \includegraphics[width=0.2\linewidth]{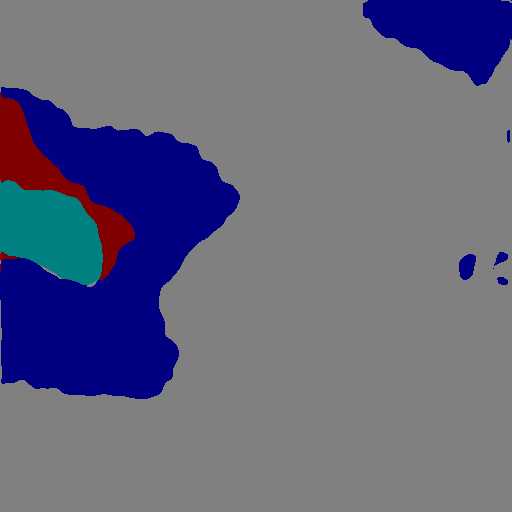}&
            \includegraphics[width=0.2\linewidth]{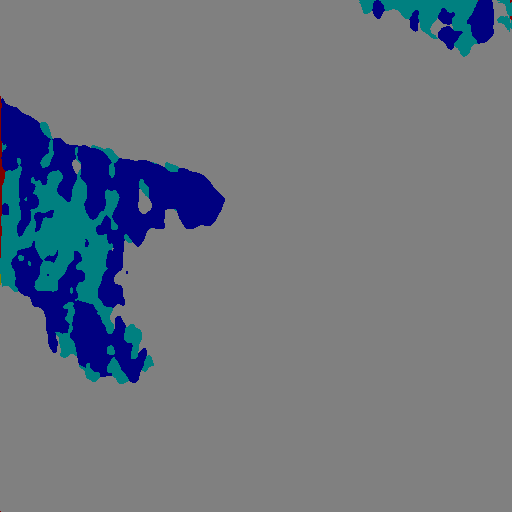}\\
            \includegraphics[width=0.2\linewidth]{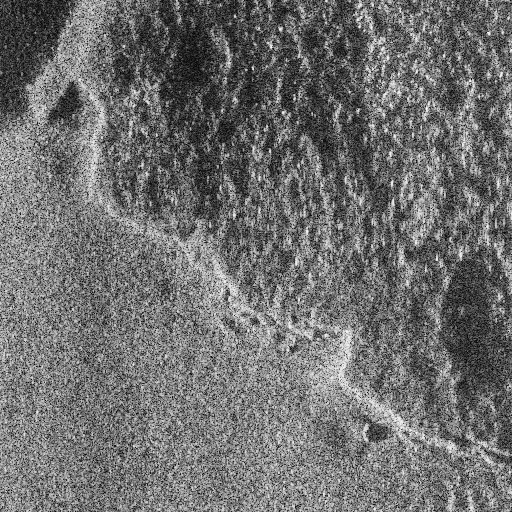}& \includegraphics[width=0.2\linewidth]{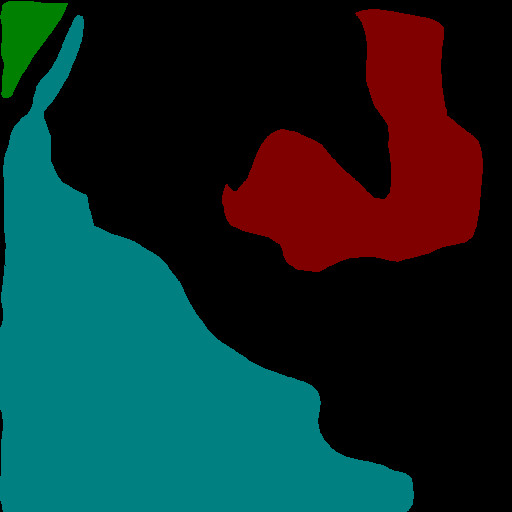}&
            \includegraphics[width=0.2\linewidth]{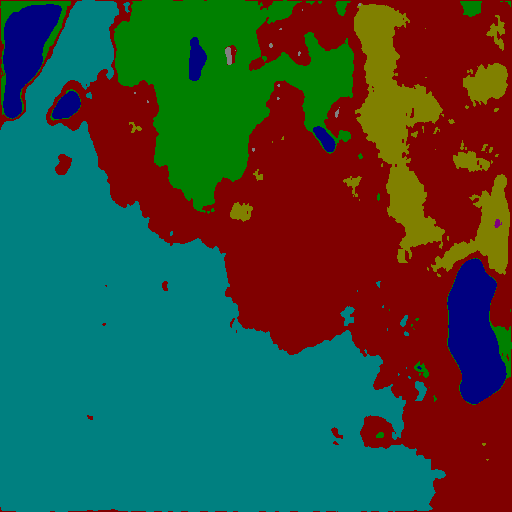}&
            \includegraphics[width=0.2\linewidth]{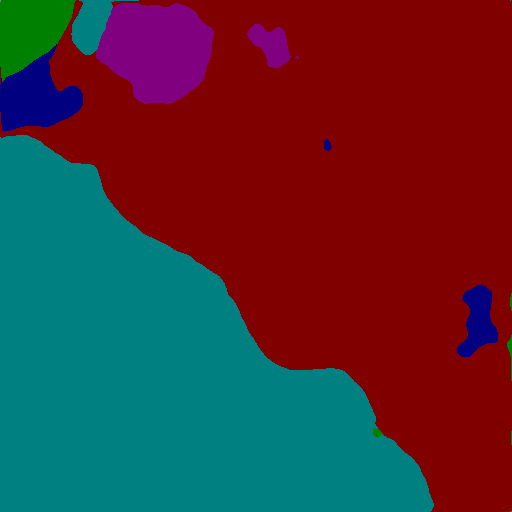}&
            \includegraphics[width=0.2\linewidth]{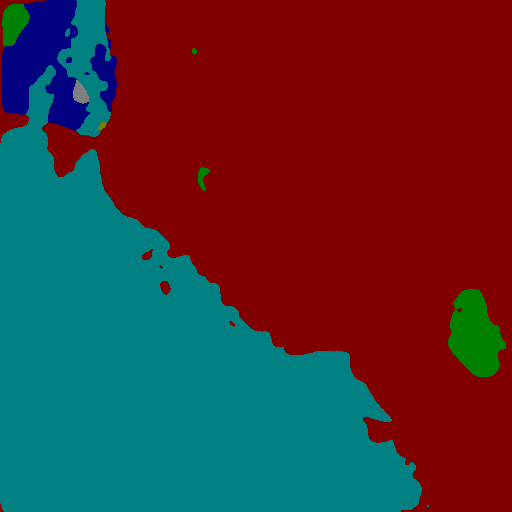}\\
            (a) Input Image  &(b) Ground Truth Labels  &(c) U-Net Result &(d) IDUS Result &(e) IDSS Result
        \end{tabular}
        \caption{The segmentation maps of seabed textures from our semi-labeled dataset.  Class-Color mapping: SH (blue), DS (green), BS (cyan), SG (red), RK (purple), SR (gold), LR (gray) and Unlabeled (black). Although IDUS is trained in an unsupervised fashion, the predictions of IDUS, IDSS and the supervised U-Net are almost the same for the ground-truth labeled regions. Especially for the bottom SAS image, we believe that the IDUS and IDSS are more consistent with the ground truth for SG (red) class and have less local heterogeneity.}
        \label{fig:visualize}  
        \vspace{-0.5cm}
\end{figure*}
\vspace{-8mm}
\subsection{Visualization of Select Results}\label{visual}
To visualize network predictions, we use the same train-test splits from Section \ref{feature_quality} to train a supervised U-Net as in the previous subsection \ref{idssevl} and IDUS. We use a subset of fifteen images of the training set to train IDSS.

The prediction results of IDUS, IDSS and the U-Net on the test set are shown in \figurename~\ref{fig:visualize}. In the top row of \figurename~\ref{fig:visualize}, although IDUS is trained in an unsupervised fashion, the predictions of IDUS, semi-supervised IDSS, and supervised U-Net are almost the same in the labeled region, especially for the Large Sand Ripple (LR). In the bottom row of \figurename~\ref{fig:visualize}, IDUS and IDSS provide more consistent predictions for the unlabeled regions and exhibit homogeneity in labeling similar textures. We see that the U-Net's predictions do not have these nice properties especially in unlabeled regions. We posit a possible reason for this: the U-Net overfits since the data is partially and sparsely-labeled. IDUS and IDSS avoid this pitfall by combining {\em domain inspired} semantic information with learning from labeled pixels. Specifically,  the superpixels in IDUS/IDSS exploit custom feature initialization to group local cluster textures having similar spatial characteristics into a single class assignment thereby mitigating over-fitting.  \figurename~\ref{fig:visualize} shows IDUS and IDSS segment SAS image regions more agreeably with the ground truth than the U-Net.

\vspace{-0.25cm}
\section{Conclusion}
We propose an unsupervised method for SAS seabed image segmentation which works by iteratively clustering on superpixels while training a deep model. Our proposed method, Iterative Deep and Unsupervised Segmentation (IDUS), does not require large amounts of labeled training data which is difficult to obtain for SAS. We demonstrate that IDUS obtains state-of-the-art performance on a real-world SAS dataset and show performance benefits by comparing against existing methods.   Additionally, we develop  an extension of IDUS called IDSS that is able to work in a weakly or semi-supervised manner. IDSS is also evaluated against supervised state of the art methods and shown to yield superior results.  Finally, we show inference (run) times for IDUS/IDSS which indicate that it can easily be deployed for real-time processing onboard low size, weight, and power (SWaP) systems. 
\section{Acknowledgments}
The authors thank Dr. J. Tory Cobb of the Naval Surface Warfare Center for providing the data used in this work.

% ==================================================================================================
% References
% ==================================================================================================
\bibliographystyle{IEEEtran}
\bibliography{main}

\end{document}